\documentclass[journal]{IEEEtran}
\usepackage{dtk-logos} 
\usepackage[utf8]{inputenc}
\usepackage{times}
\usepackage[english]{babel}
\usepackage[utf8]{inputenc}
\usepackage{amsmath, amsthm}
\usepackage{graphicx}
\usepackage{epsfig}
\usepackage[colorinlistoftodos]{todonotes}
\usepackage[hidelinks]{hyperref}
\usepackage{tikz}
\usepackage[ruled,linesnumbered]{algorithm2e}
\usepackage{amssymb}
\usepackage{nccmath}
\usepackage{mathtools}
\usepackage{bm}
\usepackage{soul}
\usepackage{mdframed}
\usepackage{algorithmic}
\usepackage{float}
\usepackage[caption = false]{subfig}
\usepackage{enumitem}

\DeclarePairedDelimiterX{\norm}[1]{\lVert}{\rVert}{#1}

\newtheorem{definition}{Definition}
\newtheorem{problem}{Problem}

\definecolor{red}{cmyk}{0.1,5,0.1,0.1}

\newcommand{\cD}{\mathcal{D}}
\newcommand{\bd}{\bm{d}}
\newcommand{\bl}{\bm{l}}
\newcommand{\bell}{\bm{\ell}}
\newcommand{\bdelta}{\bm{\delta}}
\newcommand{\btau}{\bm{\tau}}
\newcommand{\cF}{\mathcal{F}}
\newcommand{\cP}{\mathcal{P}}
\newcommand{\cQ}{\mathcal{Q}}
\newcommand{\cA}{\mathcal{A}}
\newcommand{\cS}{\mathcal{S}}
\newcommand{\cO}{\mathcal{O}}
\newcommand{\cE}{\mathcal{E}}
\newcommand{\cT}{\mathcal{T}}

\begin{document}

	\title{Topological Sweep for Multi-Target Detection of Geostationary Space Objects}

	\author{Daqi Liu, Bo Chen, Tat-Jun Chin and Mark Rutten
		\IEEEcompsocitemizethanks{\IEEEcompsocthanksitem D. Liu, B. Chen and T.-J. Chin were with School of Computer Science, The University of Adelaide. M. Rutten was with InTrack Solutions.}}


	\maketitle

	\begin{abstract}
		Conducting surveillance of the geocenric orbits is a key task towards achieving space situational awareness (SSA). Our work focuses on the optical detection of man-made objects (e.g., satellites, space debris) in Geostationary orbit (GEO), which is home to major space assets such as telecommunications and Earth observing satellites. GEO object detection is challenging due to the distance of the targets, which appear as small dim point-like objects among a background of streak-like objects. In this paper, we propose a novel multi-target detection technique based on topological sweep, to find GEO objects from a short sequence of optical images. Our topological sweep technique exploits the geometric duality that underpins the approximately linear trajectory of target objects across the sequence, to extract the targets from significant clutter and noise. Unlike standard multi-target methods, our algorithm deterministically solves a combinatorial problem to ensure high-recall rates without requiring accurate initializations. The usage of geometric duality also yields an algorithm that is computationally efficient and suitable for online processing.
	\end{abstract}


	\IEEEpeerreviewmaketitle

	\section{Introduction}

	\IEEEPARstart {V}{irtually} all aspects of modern life depend on space technology, including communications, media, commerce, and navigation. Enabling space technology are the thousands of space assets (satellites, space station, etc.)~currently in orbit, which amount to trillions of dollars of investment. With space usage projected to increase rapidly~\cite[Fig.~8]{kennewell2013overview}, in part due to the participation of new state and private operators, the number of space assets will also grow quickly.

	Greater space usage naturally leads to more ``crowding'' of the geocentric orbits by resident space objects (RSOs); these include the space assets that directly support the intended applications, as well as the orbital debris that occur as by-products of related space activities (e.g., launching, decommissioning or destruction of space assets)~\cite{Schildknecht}.
	The increase in RSOs raises the potential of collision between space assets and debris~\cite[Sec.~6.3]{kennewell2013overview}, and this has been identified as a pressing issue.

	Achieving SSA is crucial towards alleviating the risk of space asset destruction due to collisions. Broadly, SSA entails building and maintaining an up-to-date understanding of the near space environment and the contents therein~\cite{kennewell2013overview}, to enable conjunction analysis and collision prevention strategies. A key step towards achieving SSA is the detection of known, unknown and new RSOs, which can be achieved using a variety of paradigms (e.g., ground-based radar~\cite{ender2011radar,wilden2016gestra}, ground-based telescopes~\cite{yanagisawa2009activities,vsara2013ransacing}, satellite-based observers~\cite{Flohrer,priest18}) that have complementary strengths. For example, space-based detectors are more difficult to establish, but they are not as limited by weather and atmospheric effects as ground-based detectors. Realistically, a holistic SSA solution will be a combination of different approaches.




	\subsection{Our setting}\label{sec:setting}

\begin{figure*}
		\includegraphics[width=\textwidth]{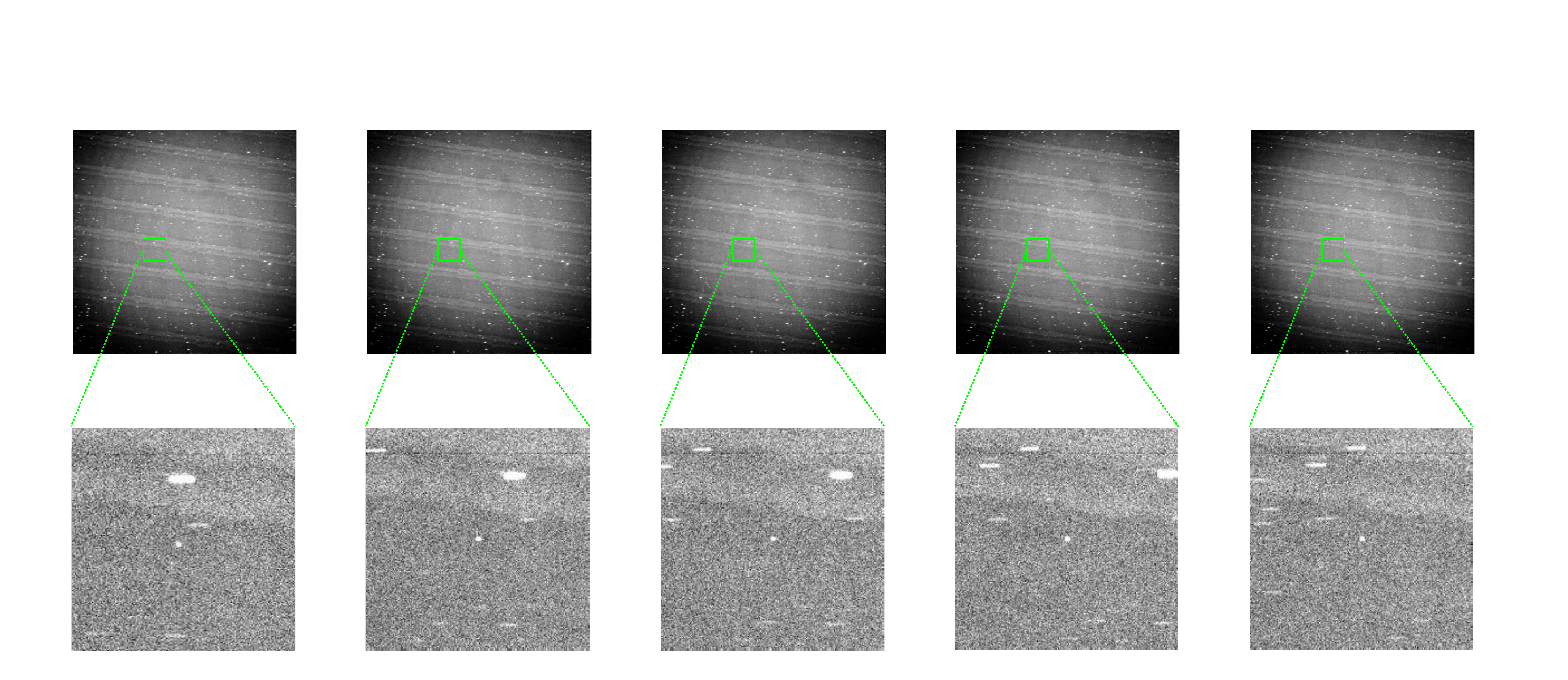}
		\caption{(Top row) Sample image sequence recorded under the setting in Sec.~\ref{sec:setting}. Here, each image is of size $2048\times2048$ pixels. (Bottom row) Close-up of a target object (an RSO in GEO) in the image sequence. The tiny size of the object (a small blob spanning a few pixels) and its relative dimness to noise and other bright celestial objects (e.g., stars which appear as streaks due to the long exposure) make detection a challenging problem.}
		\label{fig:f1}
	\end{figure*}

	In our work, we focus on the detection of RSOs in GEO, which is about $36,000$ km above equator. Objects in GEO travel synchronously with the Earth's rotation, thus they appear motionless from a fixed point on Earth. For commercial and other reasons for GEO surveillance, see~\cite{do2019robust}. We employ an optical sensor with suitable telescopic magnification to observe target regions in GEO; see~\cite{do2019robust} for specific hardware information. The optical sensor (FLI Proline PL4240 CCD array) has an angular pixel size of $\approx 4.47$ arc secs. Based on the distance between the Earth's surface and GEO, the corresponding arc length at GEO is
	\begin{equation*}
	36\times 10^{6}\times\dfrac{4.47}{3600}\times \dfrac{\pi}{180} = 780\ \text{m},
	\end{equation*}
	which is much larger than a typical GEO object. However, due to atmospheric distortion and the long exposure time (30 seconds per image), the received light from the object is smeared over a few pixels. On the other hands, Since the rotation speed of Earth is $\approx15$ arcseconds per second, the length of the streak-like stars will be $30\text{ sec} \times 15\text{ arcsec/sec} \div 4.47\text{ arcsec/pixel}\approx100$ pixels. Nonetheless, object detection is still challenging, especially against a bright star field with streak-like objects in the background; see Fig.~\ref{fig:f1} for sample images. To help deal with the paucity of the signal, following~\cite{yanagisawa2009activities,vsara2013ransacing,davey2015track,do2019robust} a short sequence of images (e.g., $5$ in total) with long exposure and short inter-image delay is acquired while fixating the camera at the target GEO region; again, see Fig.~\ref{fig:f1} for a sample sequence.

	Note that due to the roughly static position of GEO objects relative to the ground-based observer, they tend to not ``streak'' even under the long exposure. Our problem is thus reduced to finding dim point-like targets in a cluttered image sequence.

	\subsection{Existing methods for GEO object detection}

	A number of existing methods for the setting above take advantage of the approximately linear pattern of the target trajectories across the sequence, \emph{after} factoring out the apparent movement of the streak-like objects in the background. With the aid of FPGA acceleration, Yanagisawa et al.~\cite{yanagisawa2009activities,Yanagisawa2012} exhaustively search all linear trajectories across the input sequence. However, this is resource intensive and is not very attractive for space-based platforms~\cite{Flohrer,priest18}. \v{S}\'{a}ra et al.~\cite{vsara2013ransacing} and Do et al.~\cite{do2019robust} first register the images to a common image frame, then perform line finding using randomised heuristics to detect linear tracks. We adopt the framework of~\cite{vsara2013ransacing,do2019robust}, but significantly improve the track extraction step by a novel deterministic topological sweep technique~\cite{edelsbrunner1989topologically}.

	GEO object detection is also amenable to a track-before-detect (TBD) treatment~\cite{davey2013track}. The concept of TBD is to improve the SNR of weak targets by accumulating measurements across different time steps to yield more confident detections. Davey et al.~\cite{davey2015track,davey2018track} developed a histogram probabilistic multi-hypothesis tracking (H-PMHT~\cite{streit1995probabilistic}) technique for space object detection, under a similar setting as ours. However, their technique requires a relatively long image sequence to achieve sufficient accumulation, whereas the input sequence for our technique can be short (e.g., $5$ images only). Another weakness of H-PMHT is the need for accurate track initializations. We will show how exploiting the approximately linear shape of the trajectory using topological sweep helps to overcome these issues.

	\subsection{Our contributions}

	We propose a novel algorithm based on \emph{topological sweep} for detecting multiple GEO objects for the setting described in Sec.~\ref{sec:setting}. The core idea is to exploit the fundamental geometric duality of linear point tracks to enable deterministic search over all possible candidate targets. Notwithstanding the enumerative nature of the method, the usage of topological sweep -- a classical technique from computational geometry~\cite{edelsbrunner1989topologically} -- enables high processing speeds on practical input sizes.

	Moreover, unlike~\cite{vsara2013ransacing} and~\cite{do2019robust} which employ randomised heuristics for the equivalent step in the pipeline, our technique deterministically examines all possible candidates, and thus does not run the risk of missing targets. Compared to~\cite{davey2015track}, our technique is viable even for short image sequences and is not dependent on accurate track initializations. We will experimentally benchmark against the previous methods above, as well as other multi-target tracking approaches~\cite{vo15}.

	The rest of the paper is organized as follows: in Sec.~\ref{sec:prelim}, we state the necessary preprocessing, define our problem and survey previous methods that are applicable. Sec.~\ref{sec:objective} describes the mathematical formulation adopted and baseline methods. Sec.~\ref{sec:duality} inspects the dual form of the formulation before the proposed algorithm is presented in Sec.~\ref{sec:proposed}. In Sec.~\ref{sec:results}, we evaluate and compare our method against the alternatives, before concluding and mentioning future work in Sec.~\ref{sec:conclusions}.

	\section{Preliminaries}\label{sec:prelim}

	We begin by describing the preprocessing conducted in our pipeline, and a formal statement of our problem.

	\subsection{Preprocessing}\label{sec:preprocessing}


	We apply the preprocessing method of~\cite{do2019robust,vsara2013ransacing}, which takes as input a sequence of $F$ images $\{I_1,\dots,I_F\}$ (e.g., the sequence in Fig.~\ref{fig:f1}) and outputs a 2D point set
	\begin{align}
	\mathcal{D} = \{\bm{d}_i\}_{i=1}^N,
	\end{align}
	where each $\bm{d}_i = (x_i,y_i)$ has a time index
	\begin{align}
	t_i \in \{1,\dots,F\}
	\end{align}
	that indicates the image origin of $\bm{d}_i$; see Fig.~\ref{fig:f3b} for a sample result of the preprocessing. The main steps of the preprocessing are to reduce each image to a set of discrete foreground points (stars and RSOs), then align the point sets from all images onto a common reference frame.

	We apply the foreground segmentation method of~\cite{do2019robust} based on Gaussian Process regression. The aim of this step is to reduce the computational burden of the subsequent processing, by retaining only the pixel locations that matter (i.e., those corresponding to stars and RSOs). The procedure can be viewed as a statistically justified form of thresholding, which takes into account local intensity information and image structure. Nonetheless, the segmentation is conservative, in that significant false positives remain; see Fig.~\ref{fig:f3b} and ~\ref{fig:select}.

	Point set alignment is achieved by matching the star field patterns between the images, which also allow the background stars to be removed from the output since the stars overlap in the common frame; see~\cite{do2019robust} for details. The minimum number of points required for homography transformation is $4$, and hence we need at least $4$ stars to perform image alignment. However, more stars are required to improve robustness and minimize transformation error. Note that inevitable inaccuracies in the alignment will cause points corresponding to some background stars or parts thereof to remain in $\mathcal{D}$.

	In summary, with the foreground/background subtraction algorithm (Gaussian Process), noise and cloud can be  significantly reduced to improve Signal-to-Noise Ratio (SNR). We then use image alignment (homography) to remove background stars, to obtain the time-indexed 2D point set $\mathcal{D}$.
	For details of the preprocessing, see~\cite{do2019robust}. We emphasize that the preprocessing is imperfect, in that significant clutter remains in the output, alongside the target objects. Also, note that the real targets can also be occluded due to background stars, cloud cover, or significant imaging artefacts; see Fig.~\ref{fig:cover}

	\subsection{Overall aim}\label{sec:overaim}

	\begin{figure}
		\subfloat[]{\label{fig:f3b}\includegraphics[width = \linewidth]{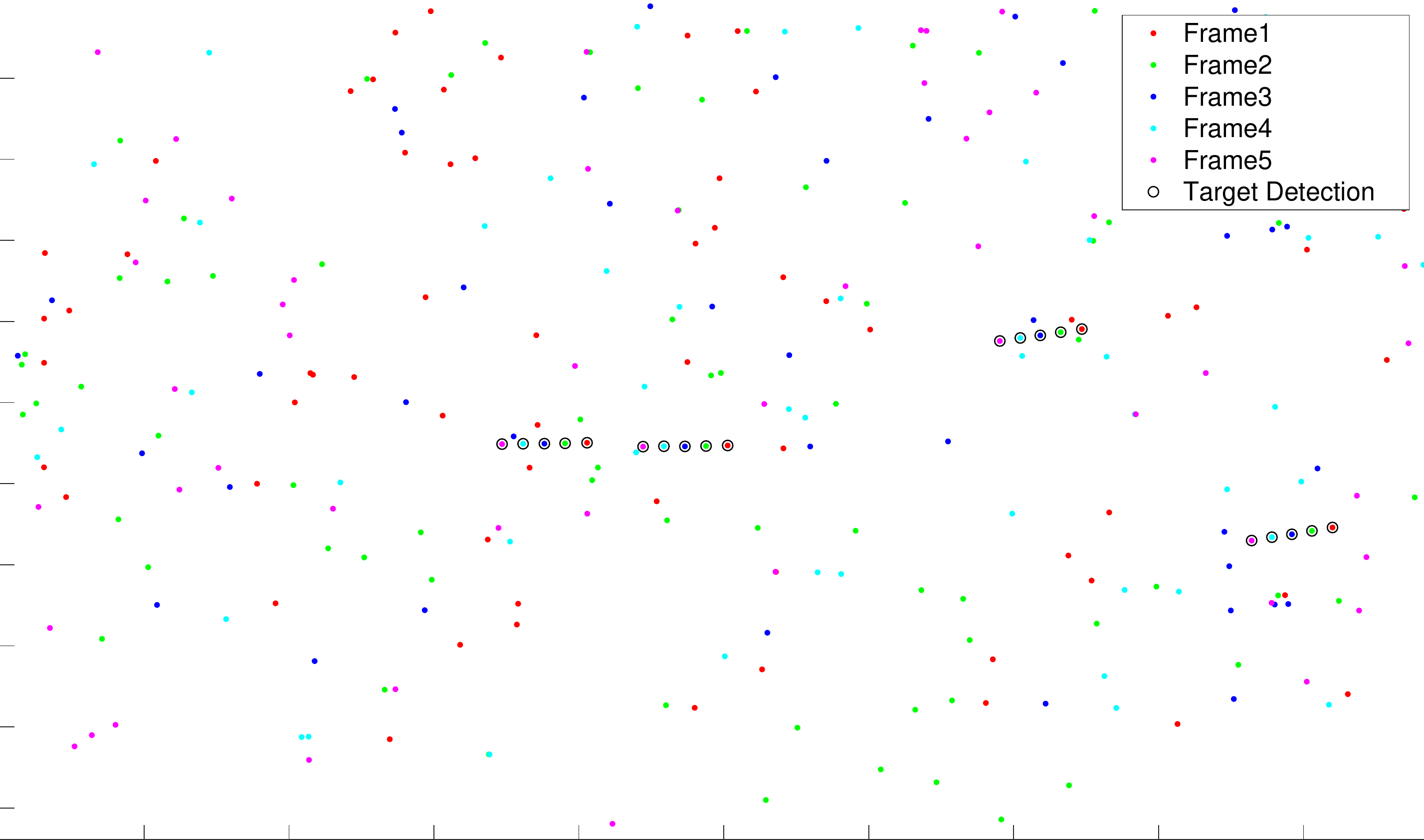}}\\
		\subfloat[]{\label{fig:f3a}\includegraphics[width = \linewidth]{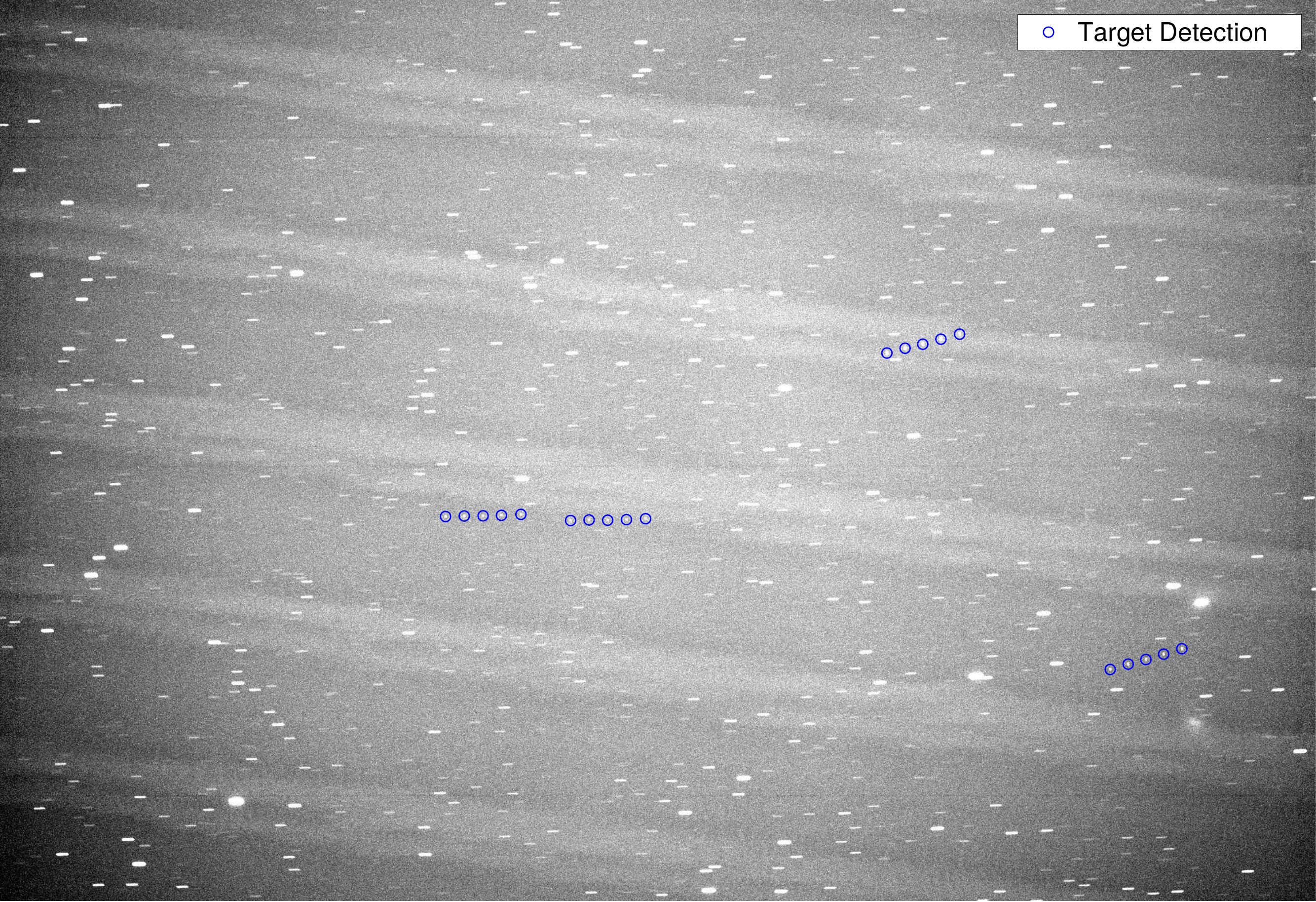}}
		\caption{(a) Time-indexed 2D points $\cD$ produced by the preprocessing of~\cite{do2019robust} on the sequence in Fig.~\ref{fig:f1}. The points are colored according to their time index. Points circled in black indicate the target objects; there are four distinct objects in this sequence, which make up four tracks $\{ \btau_k \}_{k=1}^4$. (b) Target objects plotted on the original images warped to a common frame using the registration parameters estimated during preprocessing (note that warping the images is purely for visualization and is not required in our method).}
		\label{fig:f3}
	\end{figure}
	\begin{figure}
		\includegraphics[width=\linewidth]{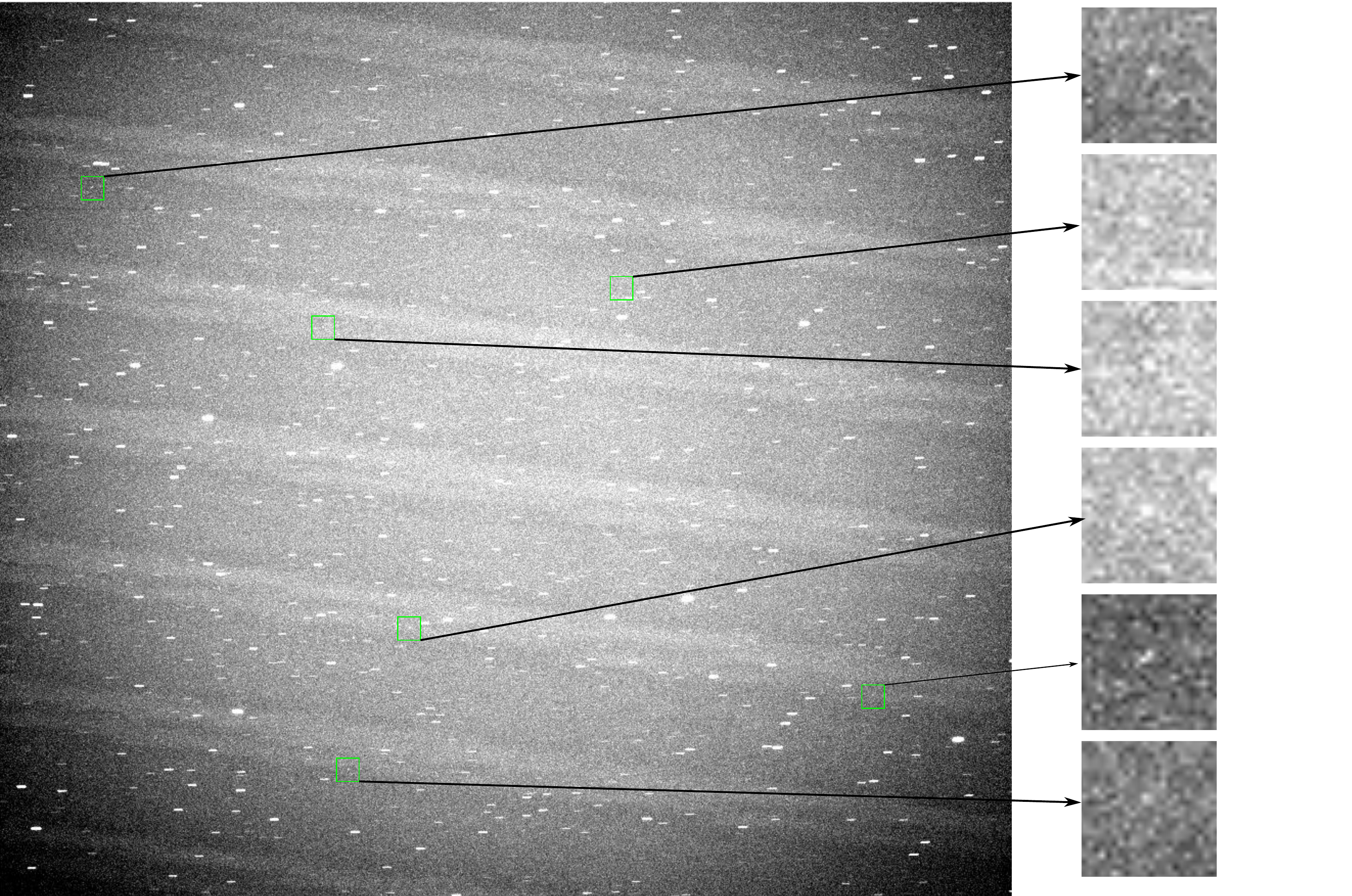}
		\caption{Many point-like objects exist in the images, due to environmental and sensor noise. In fact, in the examples above, only one is a true RSO.}
		\label{fig:select}
	\end{figure}
	\begin{figure}[ht]\centering
		\subfloat[Occlusion due to stars.]{\includegraphics[width = 0.45\linewidth]{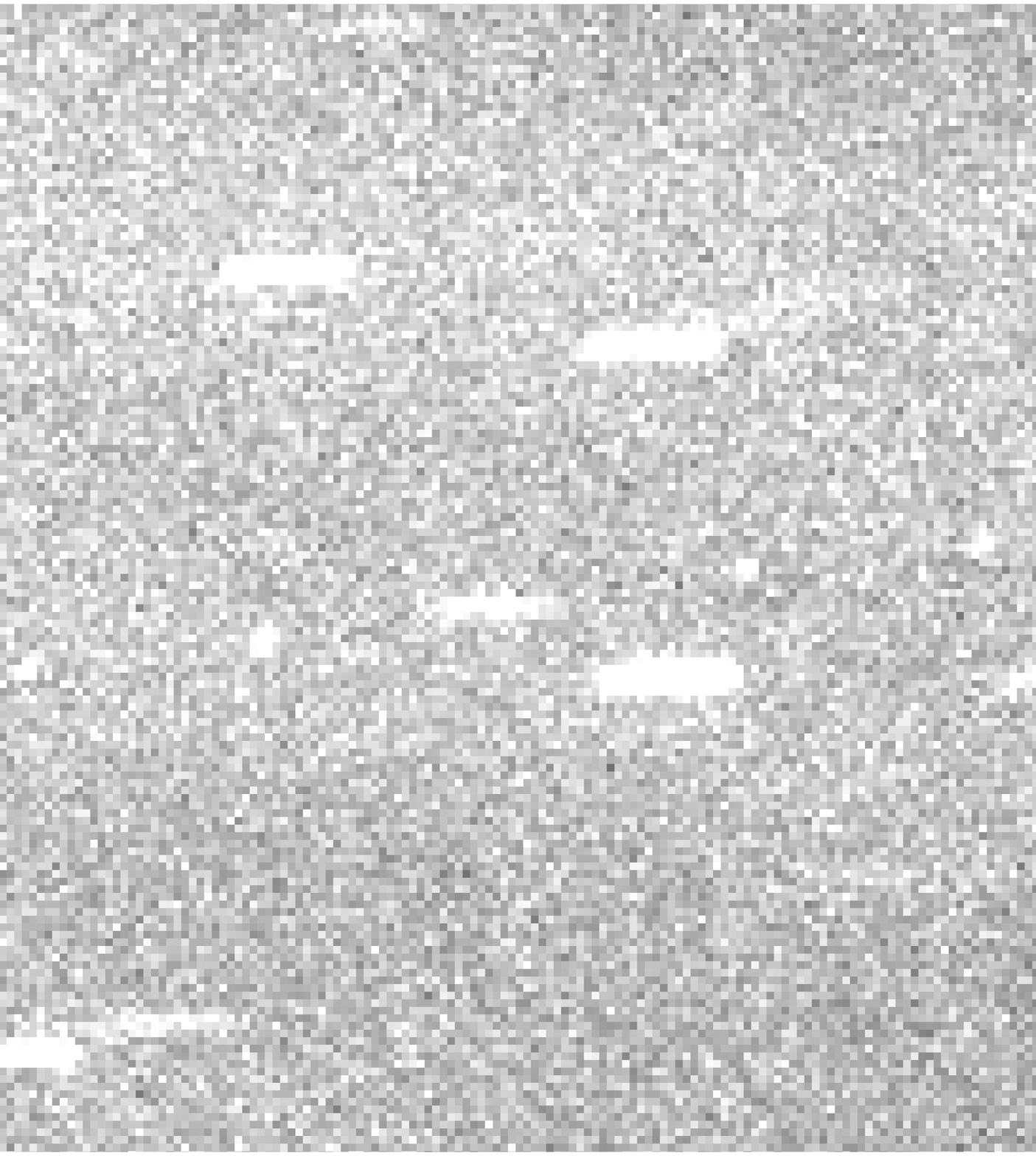}} \;\;
		\subfloat[Occlusion due to cloud cover.]{\includegraphics[width = 0.45\linewidth]{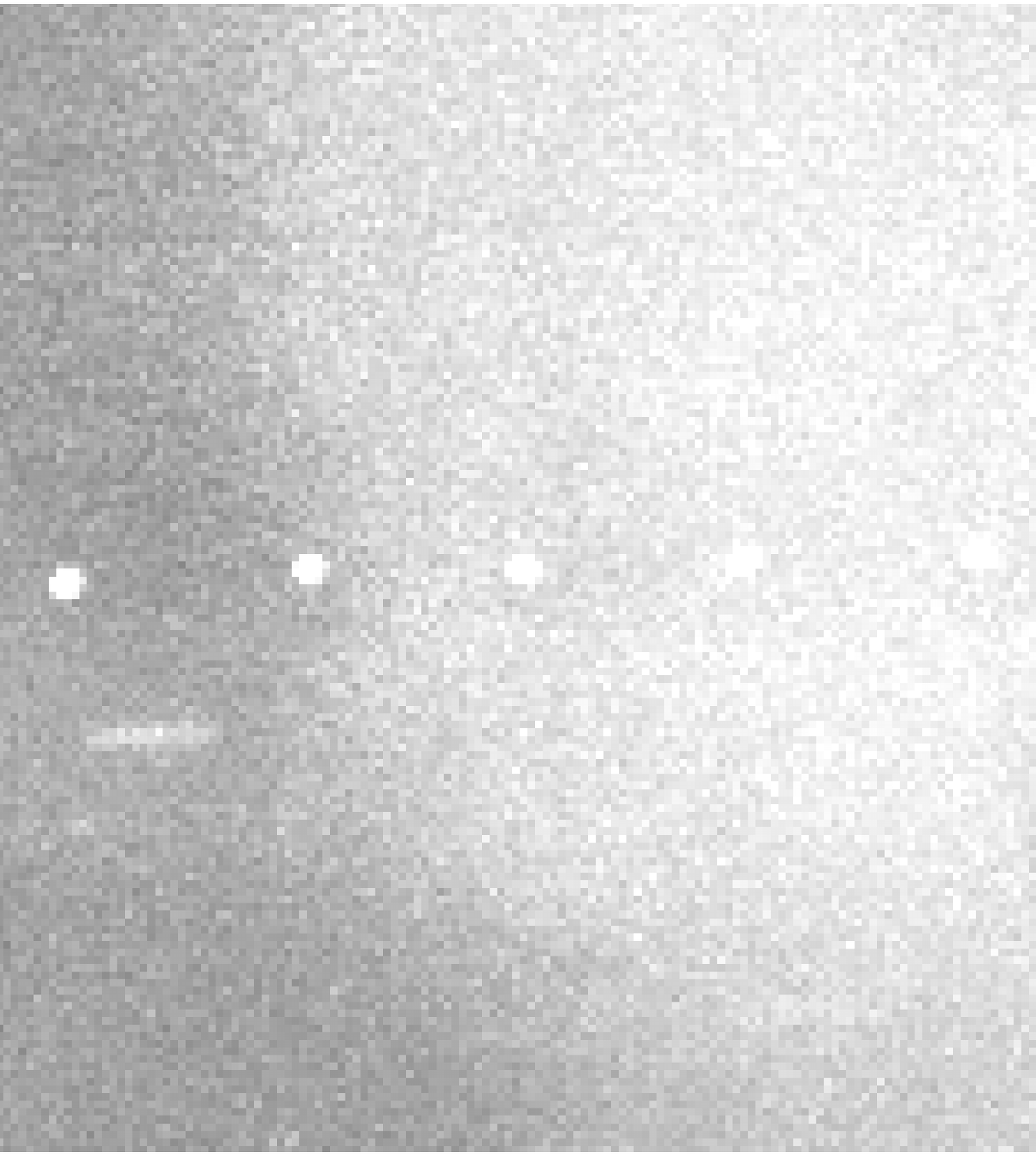}}
		\caption{Other sources of difficulties.}
		\label{fig:cover}
	\end{figure}
	Our focus in this paper starts from the time-indexed 2D point set $\mathcal{D}$. The overarching aim is to find a set of tracks
	\begin{align}\label{eq:trackdef}
	\{\btau_k\}_{k=1}^{K}
	\end{align}
	corresponding to $K$ objects, where each track
	\begin{align}
	\btau_k \subset  \mathcal{D}
	\end{align}
	contains a number of elements of $\mathcal{D}$; see Fig.~\ref{fig:f3} for the desired result. What constitutes a ``track'' and how to evaluate the quality of a track vary across different formulations (we will define ours in Sec.~\ref{sec:objective}). Also, the number of objects $K$ is potentially unknown, and the detection or tracking algorithm should be robust against this missing information.


	Some multi-target tracking algorithms do not output tracks as subsets of the input points, but the track parameters (e.g., state estimates) directly. A data association step can be performed to convert such outputs to the form~\eqref{eq:trackdef}~\cite[Chapter.~8]{mallick13}.

	\subsection{Existing methods for multi-target tracking}


	There is a large body of literature on multi-target tracking~\cite{mallick13,vo15}. Here, we highlight several techniques that are relevant to our overall aim, as defined above.

	The probabilistic multi-hypothesis tracking (PMHT) technique~\cite{rago1995comparison,willett2002pmht} can be applied on $\mathcal{D}$ to extract $K$ tracks. Briefly, given initializations to the tracks, PMHT alternates between weighted assignment of the points to the tracks, and updating the tracks based on the weighted assignments using an estimator, e.g., Kalman filter. It can be shown that the procedure is a form of hill climbing that is guaranteed to converge. However, the quality of the final output depends heavily on the goodness of the initializations~\cite{streit2006pmht}.


	By assuming that the target trajectories form linear tracks, the Hough Transform (HT) technique~\cite{moyer2011multi,sahin2014multi} can be applied to the time-indexed point set $\mathcal{D}$ to find the desired trajectories $\{ \btau_k \}^{K}_{k=1}$, which correspond to $K$ peaks in the Hough accumulator. However, the accuracy of HT is sensitive to parameter tuning; for example, the appropriate resolution of the Hough accumulator is tricky to determine \emph{a priori}. Incorrect parameter settings will cause genuine tracks to be suppressed, or false tracks corresponding to spurious peaks to be returned.

	Also under linear trajectory assumption, Guo and White~\cite{guo2015plane} applied random sample consensus (RANSAC)~\cite{fischler1981random} and \emph{plane sweep}~\cite{deberg2008,chin2017maximum} to perform TBD on point-wise measurements. They showed higher accuracy and stability than HT; however, their technique has been demonstrated only on single-target scenarios and synthetic data. Our approach is inspired by~\cite{guo2015plane}, specifically, the usage of computational geometry techniques based on the concept of duality. However, we extend~\cite{guo2015plane} to the multi-target case, improve their runtime using topological sweep~\cite{edelsbrunner1989topologically}, and validate our method on real data from SSA.

	Given the low speed of the GEO objects relative to the observer and the short length of the input sequence, the target trajectories are modeled well by lines. An additional property that we take advantage of is the nearly constant distances between adjacent target positions lying on the same track, which are due to constant exposure times and frame rates used during the image capturing. Fig.~\ref{fig:f3} demonstrates the validity of this assumption. With this assumption, our approach can handle missing detection in some frames since the remaining points can still be modeled by lines.

	\section{Problem Formulation}\label{sec:objective}

	\subsection{Objective}

	Based on the above observations, we construct a model for our targets as defined in the following. First, we concatenate $x_i$ and $t_i$ with $1$ to form vectors $[x_i~1]$ and $[t_i~1]$.
	\begin{definition}[Feasible track]\label{def:feasible}
			A track $\btau$ is feasible if
			\begin{itemize}
				\item[C1:] $\bd_i, \bd_j \in \btau \implies t_i \ne t_j$ (no two points in the track originated from the same image).
				\item[C2:] There exists $\bl_1 \in \mathbb{R}^2$ such that for all $\bd_i \in \btau$,
				\begin{align}
				\left| y_i - \left[x_i~1\right]\bl_1 \right| \le \epsilon_{1}
				\end{align}
				(all points in the track lie within distance $\epsilon_1$ to a line).
				\item[C3:] There exists $\bl_2 \in \mathbb{R}^2$ such that for all $\bd_i, \bd_j \in \btau$,
				\begin{align}
				\left| x_i - \left[t_i~1\right]\bl_2 \right| \le \epsilon_{2}
				\end{align}
				(points on the track are ordered according to their time index and separated by a constant distance $\left| x_i - \left[t_i~1\right]\bl_2 \right| $ (up to error $\epsilon_2$) along the $x$-axis).
			\end{itemize}
		\end{definition}

\begin{figure}
			\subfloat[]{\label{fig:f4a}\includegraphics[width = \linewidth/2]{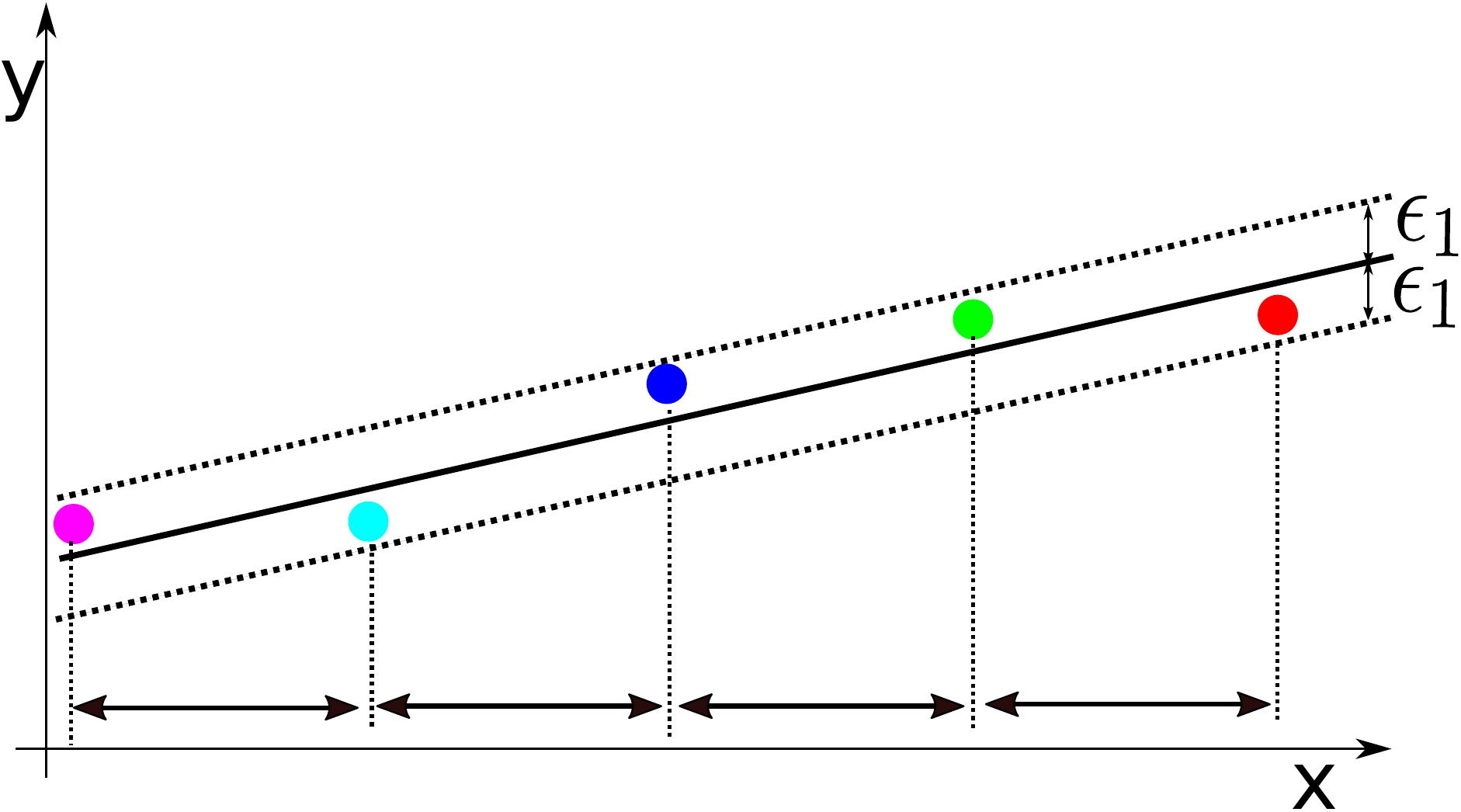}}
			\subfloat[]{\label{fig:f4d}\includegraphics[width = \linewidth/2]{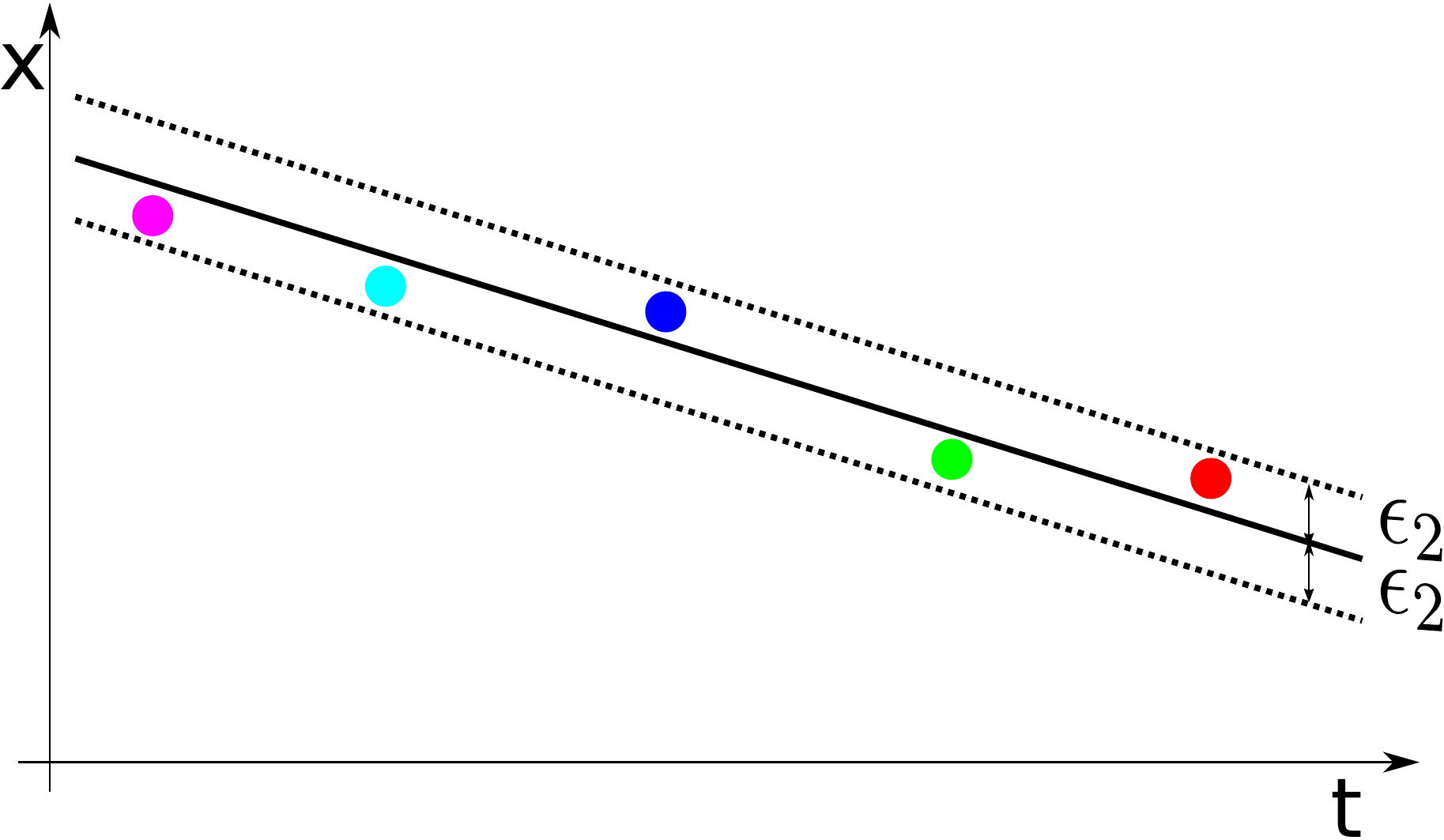}}
			\caption{(a) A feasible track consists of points from different images (C1); here, the points are color-coded as per Fig.~\ref{fig:f3b}. The points also lie within $\epsilon_1$ to a line $\bl_1$ in $(x,y)$ space (C2) and are ordered and separated by a constant distance (up to error $\epsilon_2$) along the $x$-axis (C3). (b) Condition C3 is also captured by the points lying within $\epsilon_2$ to a line $\bl_2$ in $(t,x)$ space.}
			\label{fig:f4}
		\end{figure}

\begin{figure}
			\subfloat[]{\label{fig:f5d}\includegraphics[width = \linewidth/2]{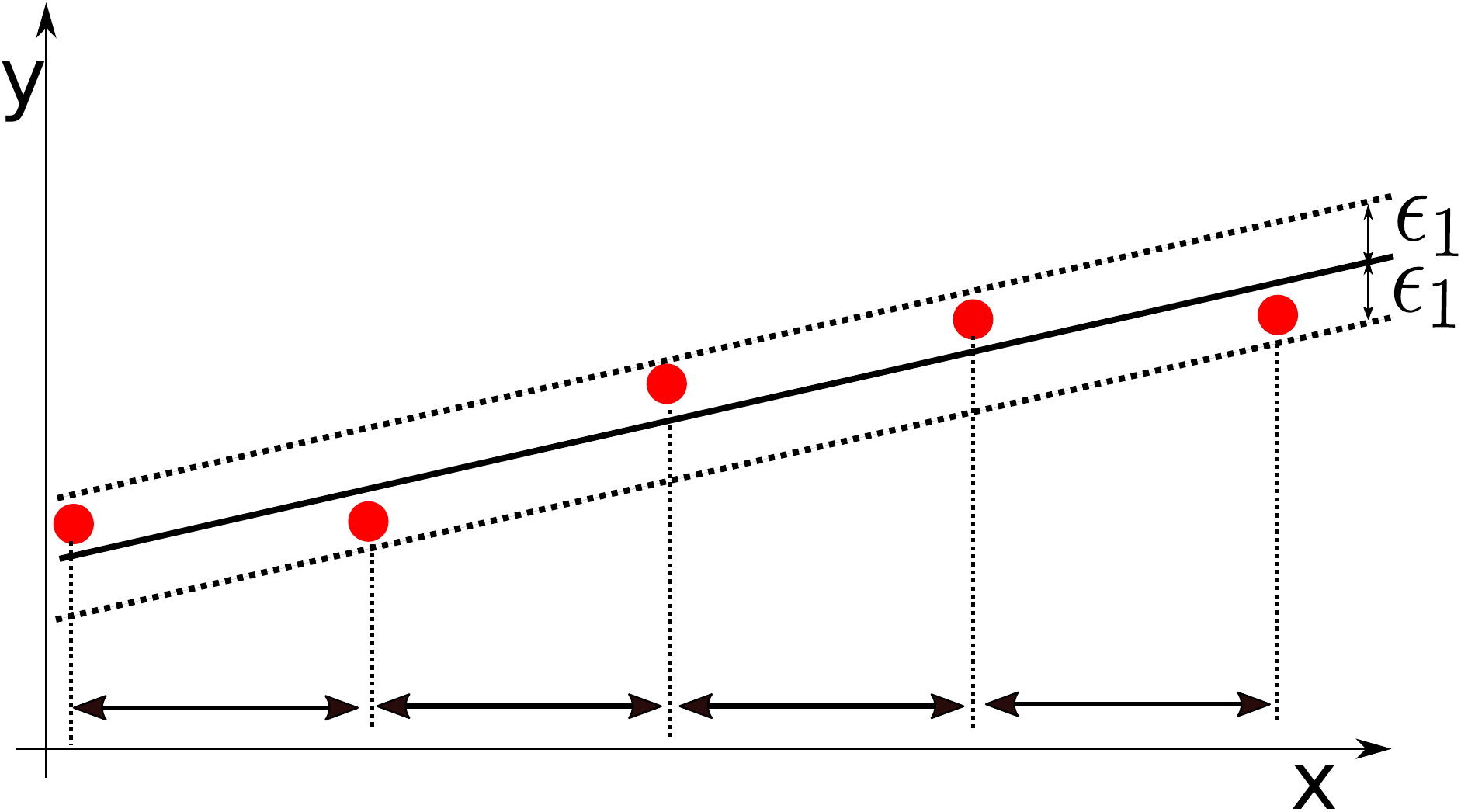}}
			\subfloat[]{\label{fig:f5a}\includegraphics[width = \linewidth/2]{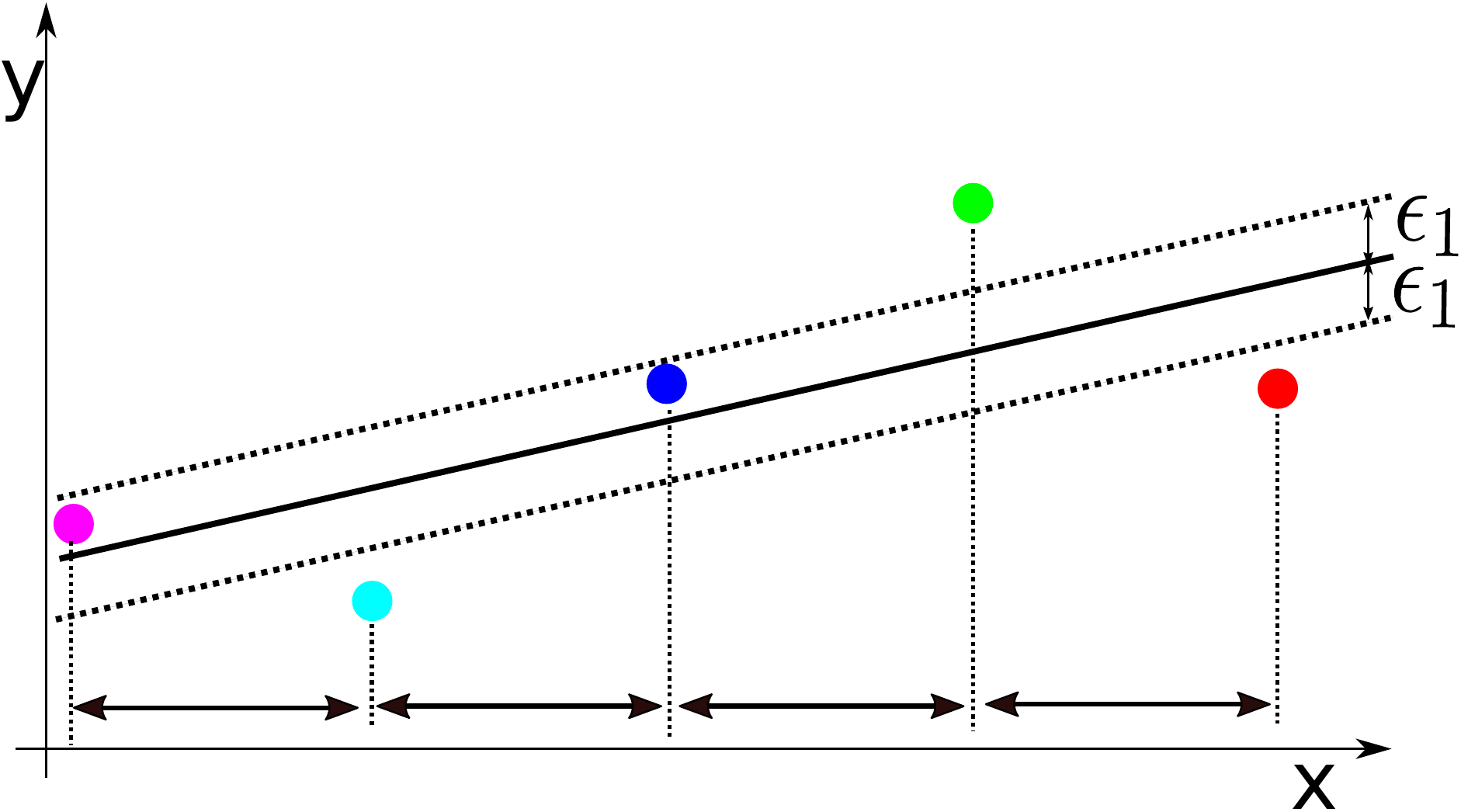}}\\
			\subfloat[]{\label{fig:f5b}\includegraphics[width = \linewidth/2]{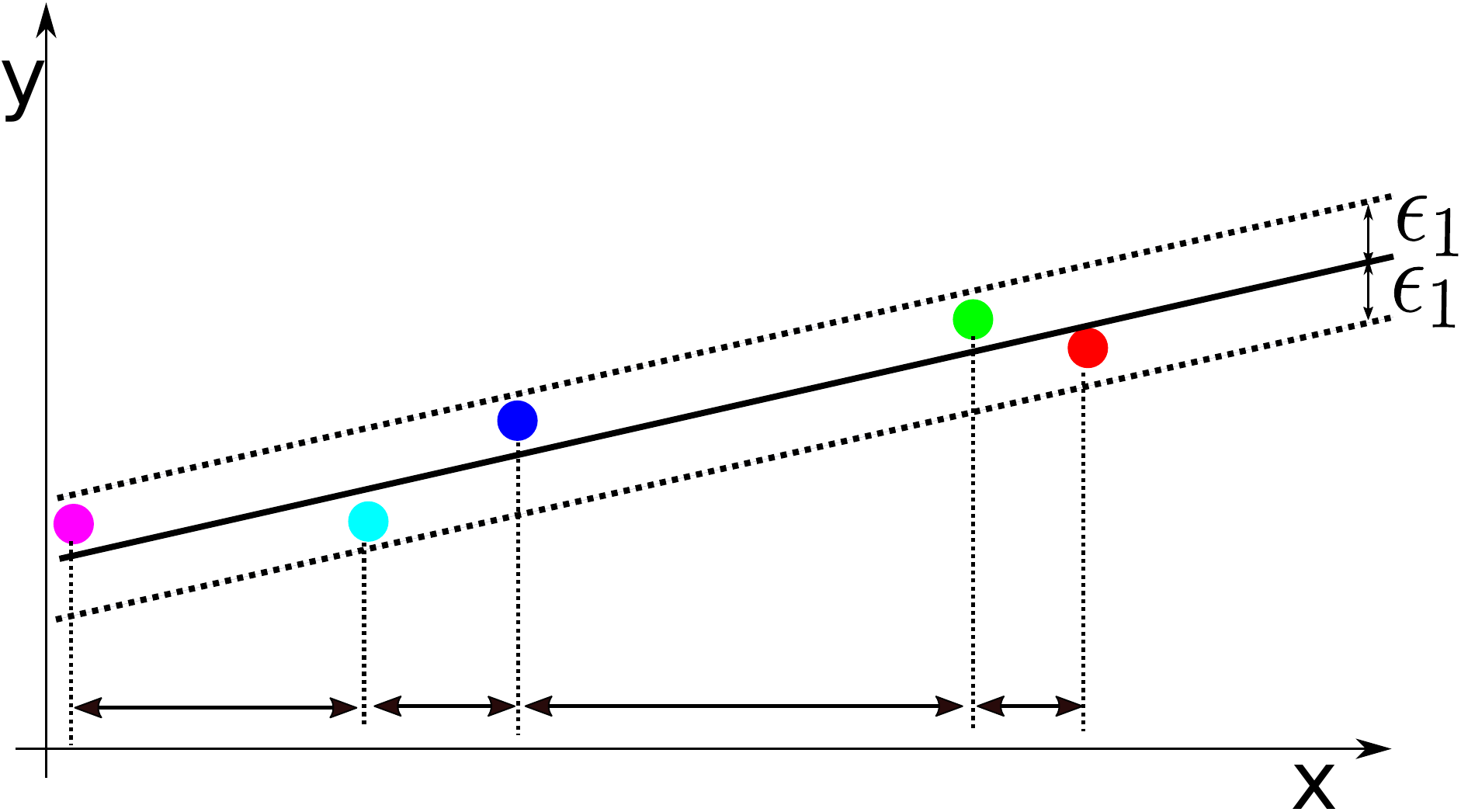}}
			\subfloat[]{\label{fig:f5c}\includegraphics[width = \linewidth/2]{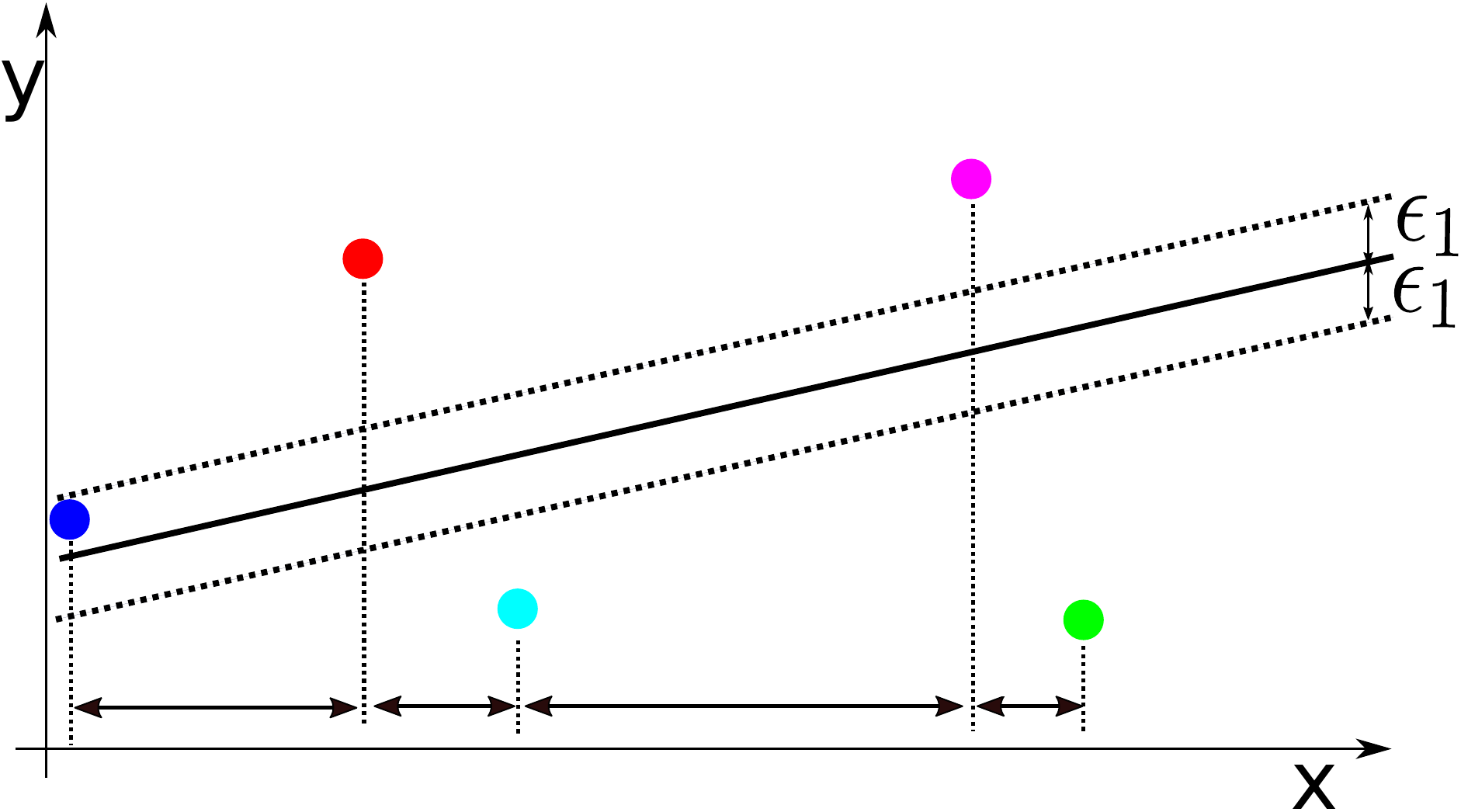}}
			\caption{Illustration of infeasible tracks. (a) The points are not from different time indices (violates C1). (b) The points do not form a line (violates C2). (c) The points are not separated by a constant distance up to error $\epsilon_2$ (violates C3). (d) The points are located randomly and meet none of the conditions.}
			\label{fig:f5}
		\end{figure}

		Figs.~\ref{fig:f4} and \ref{fig:f5} illustrate feasible and infeasible tracks. We then refine the overall aim in Sec.~\ref{sec:overaim} into the following.

		\begin{problem}[FINDALLTRACKS]\label{prob:findalltracks}
			Given a time-indexed 2D point set $\mathcal{D} = \{ \bm{d}_i \}^{N}_{i=1}$ and positive thresholds $\epsilon_1$ and $\epsilon_2$, find all feasible tracks in $\mathcal{D}$.
		\end{problem}

		As defined above, the solution to FINDALLTRACKS is the list of \emph{all} feasible tracks $\mathcal{T} = \{ \btau_1,\btau_2,\dots \}$ in $\mathcal{D}$. Note that not all feasible tracks are meaningful; in the degenerate case, any two points will satisfy C2 and C3, thus tracks of length two should be ignored. Also, due to C1, a feasible track has length at most $F$ (the number of images). To provide a result for the overall aim in Sec.~\ref{sec:overaim} based on FINDALLTRACKS, the $K$ longest tracks from $\mathcal{T}$ are designated as the final output, with ties broken using suitable heuristics; the GEO object detection result in Fig.~\ref{fig:f3} was obtained in this manner.

		A weakness of the model in Definition~\ref{def:feasible} is the potential numerical instability for tracks with a near vertical direction (when the slope component in $\bl_1$ approaches infinity\footnote{This issue does not affect $\bl_2$ since the time indices $t_i$ are discrete.}), thereby causing such tracks to be missed. In fact, tracks that are truly vertical (infinite slope) are not defined under the model. A simple trick to avoid this shortcoming is to swap the $x_i$ and $y_i$ coordinates in $\cD$ and solve FINDALLTRACKS a second time to detect near vertical tracks. Since our proposed algorithm (Sec.~\ref{sec:proposed}) is fast, this does not introduce significant overheads.

		\subsection{Naive method}

		A simple method for FINDALLTRACKS is to enumerate all subsets of $\mathcal{D}$ of size greater than two, and retain only the subsets that correspond to feasible tracks; see Algorithm~\ref{alg:naive}. To check that a candidate $\btau$ satisfies C2, we solve
		\begin{align}\label{eq:minimax}
		\begin{aligned}
		& \underset{\bl_1 \in \mathbb{R}^2}{\text{min}}
		& & \max_{\bd_i \in \btau}~|y_i - [x_i~1]\bl_1|
		\end{aligned}
		\end{align}
		and examine if the optimal value is no greater than $\epsilon_1$. The minimax problem~\eqref{eq:minimax} can be solved analytically; see~\cite[Chapter~2]{cheney00} for details. C3 can be tested by simply changing the entering measurements in the minimax problem
		\begin{align}
		\begin{aligned}
		& \underset{\bl_2 \in \mathbb{R}^2}{\text{min}}
		& & \max_{\bd_i \in \btau}~|x_i - [t_i~1]\bl_2|,
		\end{aligned}
		\end{align}
		as well as changing the comparison threshold to $\epsilon_2$.

		\begin{algorithm}[t]\label{alg:naive}
			\caption{Naive method for Problem~\ref{prob:findalltracks}.}
			\begin{algorithmic}[1]
				\REQUIRE Time-indexed 2D point set $\mathcal{D} = \{ \bm{d}_i \}^{N}_{i=1}$, positive thresholds $\epsilon_1$ and $\epsilon_2$.
				\STATE $\mathcal{T} \leftarrow \emptyset$.
				\FOR{all $\btau \subset \mathcal{D}$ that satisfy C1 and $|\btau| > 2$}
				\IF{$\btau$ satisfies C2 and C3}
				\STATE $\mathcal{T} \leftarrow \mathcal{T} \cup \{ \btau \}$.
				\ENDIF
				\ENDFOR
				\RETURN $\mathcal{T}$.
			\end{algorithmic}
		\end{algorithm}

		The naive method is inefficient due to exhaustive search. To simplify analysis, assume that each image contains $n$ points after the preprocessing i.e., $N = Fn$. We must thus examine
		\begin{align}
		n^F + n^{F-1} + \dots + n^3 \equiv \mathcal{O}(n^F)
		\end{align}
		subsets, which is impractical except for small $n$ and $F$.

		\subsection{Baseline method}

		To avoid the significant cost of subset enumeration, we can leverage the geometric constraint in C2. Specifically, using a line finding algorithm, we first extract a number of \emph{linear structures} $\{ \mathcal{D}_1, \mathcal{D}_2, \dots, \mathcal{D}_M \}$ from $\mathcal{D}$, where each $\mathcal{D}_j \subseteq \mathcal{D}$, $j \in \{1,\dots,M\}$, consists of points that lie close to a line. We then invoke Algorithm~\ref{alg:naive} on each $\mathcal{D}_j$ and accumulate the results to form $\mathcal{T}$. Algorithm~\ref{alg:baseline} summarizes the procedure. Since typically $|\cD_j| \ll |\cD|$, we avoid incurring the significant cost of running the naive method on large point sets.

		\begin{algorithm}[ht]\label{alg:baseline}
			\caption{Baseline method for Problem~\ref{prob:findalltracks}.}
			\begin{algorithmic}[1]
				\REQUIRE Time-indexed 2D point set $\mathcal{D} = \{ \bm{d}_i \}^{N}_{i=1}$, positive thresholds $\epsilon_1$ and $\epsilon_2$.
				\STATE $\{ \mathcal{D}_1, \dots, \mathcal{D}_M \} \leftarrow$ Run line finding algorithm on $\mathcal{D}$.\label{step:linefinding}
				\STATE $\mathcal{T} \leftarrow \emptyset$.
				\FOR{$j = 1,\dots,M$}
				\STATE $\mathcal{T}_j \leftarrow$ Run Algorithm~\ref{alg:naive} on $\mathcal{D}_j$, $\epsilon_1$ and $\epsilon_2$.
				\STATE $\mathcal{T} \leftarrow \mathcal{T} \cup \mathcal{T}_j$.
				\ENDFOR
				\RETURN $\mathcal{T}$.
			\end{algorithmic}
		\end{algorithm}

		Various line finding algorithms can be used in Step~\ref{step:linefinding} in Algorithm~\ref{alg:baseline}. For example, HT~\cite{hart72} can be executed on $\cD$ to return $M$ linear structures corresponding to $M$ peaks in the Hough accumulator. RANSAC~\cite{fischler1981random} is another popular algorithm for line finding. While standard RANSAC is designed to find only one structure, the algorithm can be executed sequentially by removing the inliers found at each run.

		Algorithm~\ref{alg:baseline} is a close depiction of \v{S}\'{a}ra et al.~\cite{vsara2013ransacing} and Do et al.~\cite{do2019robust}, who used RANSAC-like algorithms to extract tracks from $\cD$. However, as alluded above, both HT and RANSAC are heuristics; moreover, HT is sensitive to parameter tuning. Thus, using such algorithms in Algorithm~\ref{alg:baseline} does not guarantee solving FINDALLTRACKS, thus potentially missing valid feasible tracks; we will demonstrate this weakness in the experiments. In the following, we propose a novel method for FINDALLTRACKS using topological sweep.

		\section{Dual Formulation}\label{sec:duality}

		Before formulating our algorithm in Sec.~\ref{sec:proposed}, we describe a form of geometric duality and its implication on Definition~\ref{def:feasible}.

		\subsection{Point-and-line duality}
		\begin{figure}[t]\centering
			\subfloat[Primal space.]{\label{fig:f6a}\includegraphics[width = 0.49\columnwidth]{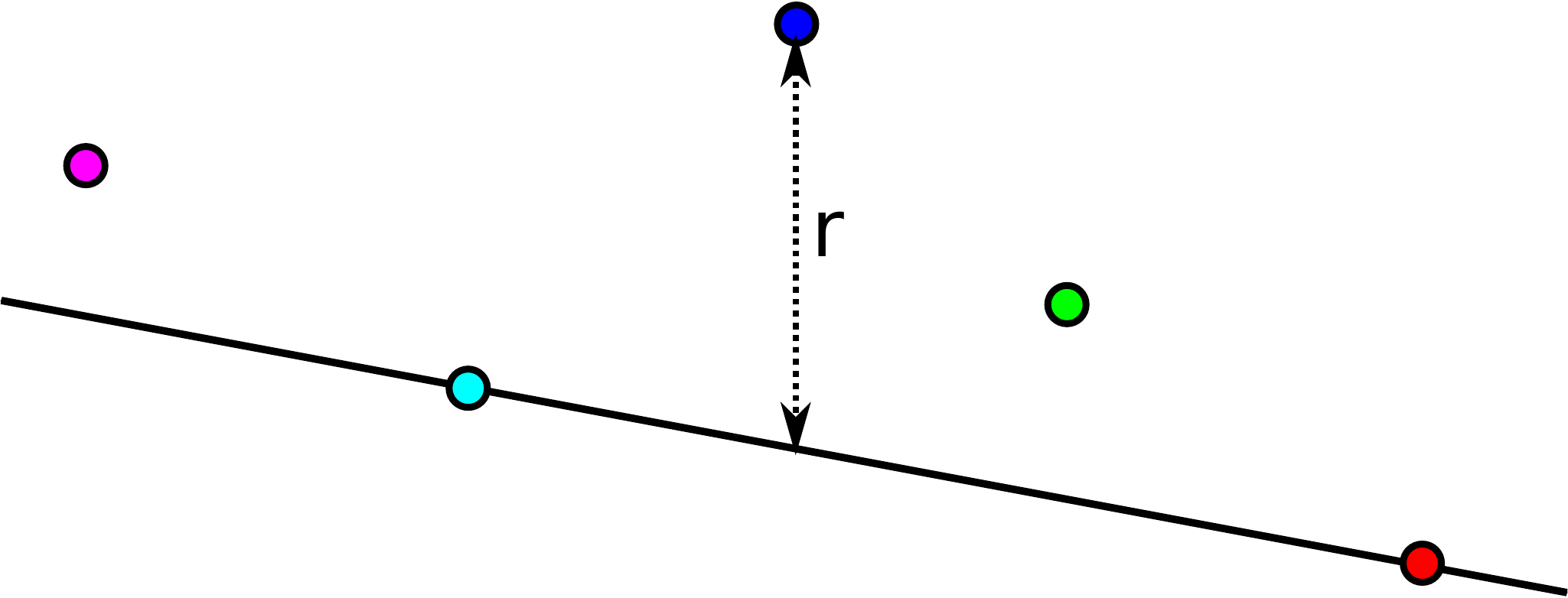}}
			\hfill
			\subfloat[Dual space.]{\label{fig:f6b}\includegraphics[width = 0.49\columnwidth]{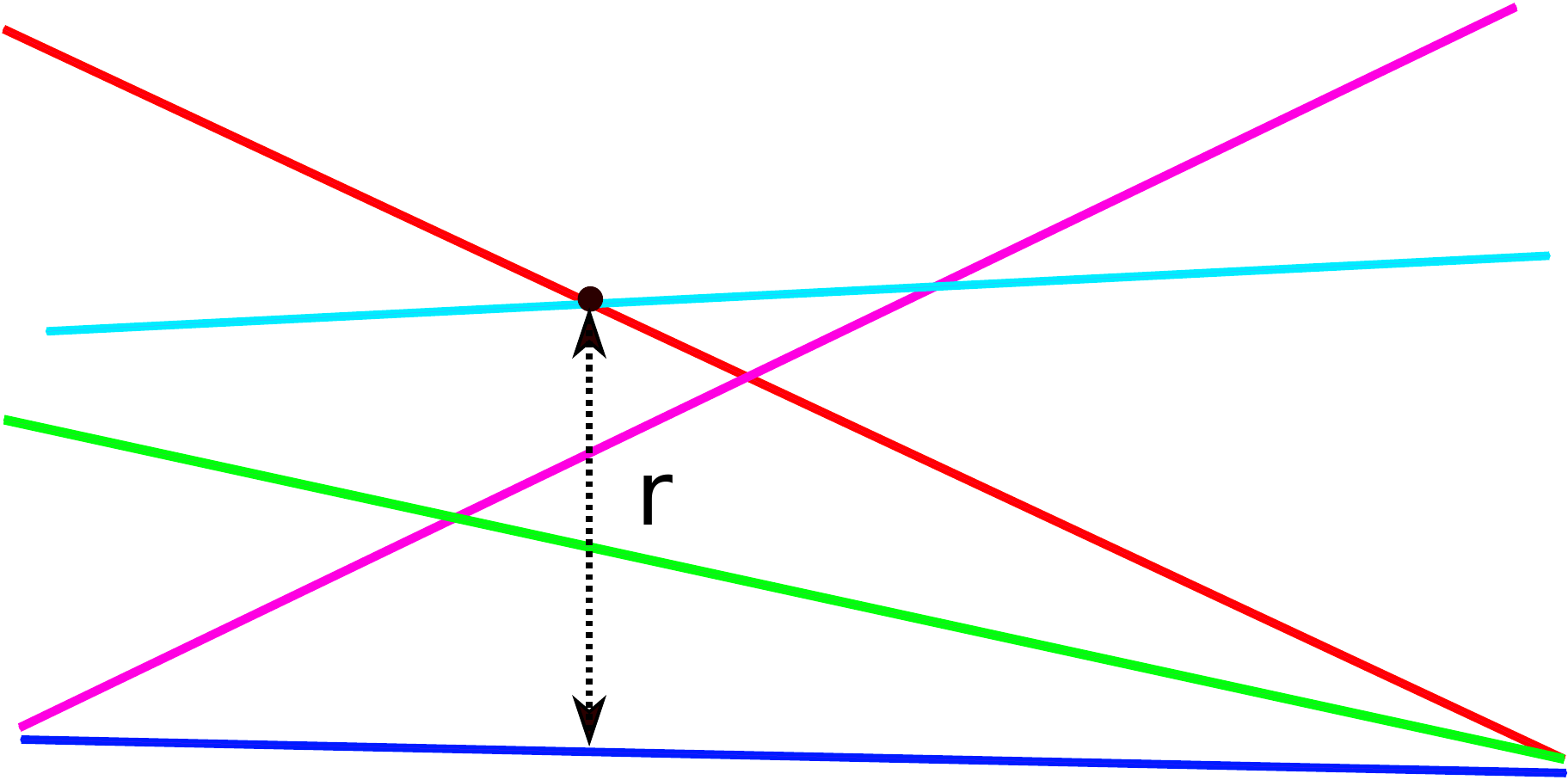}}
			\caption{Illustrating point-and-line duality. Each point in the primal space (a) is mapped to a line in the dual space (b). A line that passes through two points in the primal (e.g., cyan and red points) is mapped to the intersection of the dual lines of the two points. The distance $r$ between a point to a line in the primal (e.g., blue point to black line) is preserved in the dual; moreover, the above and below relationships are flipped between primal and dual (e.g., if a point is above a line in the primal, the dual line is below the dual point).}
			\label{fig:f6}
		\end{figure}
		We adopt the point-and-line duality originally used in~\cite{souvaine1987time}: a point $\bm{d} = (x,y)$ in the original data space $\cP$ (a.k.a.~the \emph{primal space}) is mapped to a line $\bm{\ell} = (x,y)$ in the \emph{dual space} $\cQ$; more specifically, the points $(p,q) \in \cQ$ that lie on $\bell$ are
		\begin{align}
		\{ (p,q) \in \cQ \mid q = xp + y = \left[p~1\right]\bell \}.
		\end{align}
		We summarise this primal-to-dual mapping by
		\begin{align}
		\mathcal{F}(\bd) = \bell.
		\label{eq:p2l}
		\end{align}
		$\mathcal{F}$ also maps a line $\bl = (m,c)$ in $\cP$, i.e., the set
		\begin{align}
		\{ (x,y) \in \cP \mid y = mx + c = \left[x~1\right]\bl \},
		\end{align}
		to a point $\bdelta = (-m,c)$ in $\cQ$. This is also summarised as
		\begin{align}
		\mathcal{F}(\bl) = \bdelta.
		\label{eq:l2p}
		\end{align}
		The reverse (dual-to-primal) mapping is represented as $\bd = \mathcal{F}^{-1}(\bell)$ and $\bl = \cF^{-1}(\bdelta)$. Fig.~\ref{fig:f6} illustrates this concept of duality. The implications of the adopted duality on several fundamental geometric relationships are described below.

		\subsubsection{Intersections}

		A line $\bar{\bl} = (\bar{m}, \bar{c})$ passes through two points $\bd_i$ and $\bd_j$ in $\cP$ if equations
		\begin{align}\label{eq:intersection}
		y_i = \bar{m}x_i + \bar{c} \;\;\;\; \text{and} \;\;\;\; y_j = \bar{m}x_j + \bar{c}
		\end{align}
		are satisfied simultaneously. Then, the dual point $\bar{\bdelta} = (-\bar{m},\bar{c}) = \cF(\bar{\bl})$ lies at the intersection of the lines $\bell_i = \cF(\bd_i)$ and $\bell_j = \cF(\bd_j)$, since~\eqref{eq:intersection} implies that
		\begin{align}
		q = x_ip + y_i \;\;\;\; \text{and} \;\;\;\; q = x_jp + y_j
		\end{align}
		are solved simultaneously by setting $q = \bar{c}$ and $p = -\bar{m}$. The reverse also holds: the line that passes through two points $\bdelta_i$ and $\bdelta_j$ in $\cQ$ is dual to the point that lies at the intersection of lines $\bl_i = \cF(\bdelta_i)$ and $\bl_j = \cF(\bdelta_j)$ in $\cP$. See Fig.~\ref{fig:f6}.

		\subsubsection{Point-to-line distances}

		The distance between the point $\bd_i = (x_i,y_i)$ and the line $\bar{\bl} = (\bar{m},\bar{c})$ in $\cP$ is
		\begin{align}
		r(\bd_i,\bar{\bl}) = |y_i - (\bar{m}x_i + \bar{c})| = |y_i - \left[x_i~1\right]\bar{\bl} |,
		\end{align}
		which is also the distance between the point $\bar{\bdelta} = (-\bar{m},\bar{c}) = \cF(\bar{\bl})$ and the line $\bell_i = (x_i,y_i) = \cF(\bd_i)$ and in $\cQ$
		\begin{align}
		r(\bar{\bdelta},\bell_i) = |\bar{c} - ( -x_i\bar{m} + y_i ) | = |(\bar{m}x_i + \bar{c})-y_i|.
		\end{align}
		Note that the above and below relationships are flipped, in that if $\bd_i$ is above $\bar{\bl}$, then $\cF(\bd_i)$ is above $\cF(\bar{\bl})$ and vice versa. In other words,
		\begin{align}
		\text{sign}(y_i - (\bar{m}x_i + \bar{c})) \ne \text{sign}((\bar{m}x_i + \bar{c})-y_i).
		\end{align}
		Fig.~\ref{fig:f6} also illustrates point-to-line distances under duality.



		\subsection{Feasibility conditions under duality}\label{sec:conddual}

\begin{figure}[t]\centering
			\subfloat[Primal space.]{\label{fig:f7a}\includegraphics[width = 0.49\columnwidth]{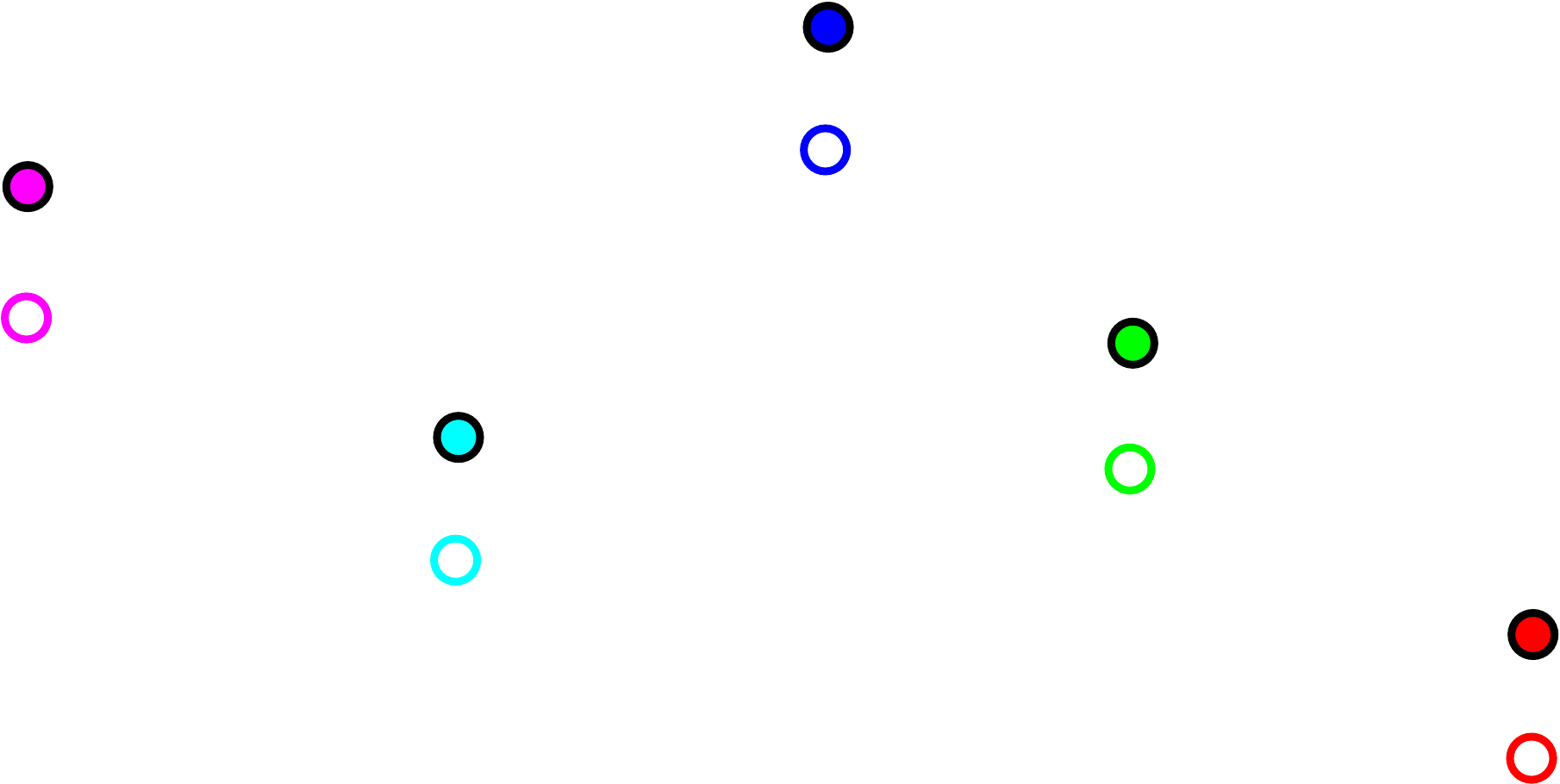}}
			\hfill
			\subfloat[Dual space.]{\label{fig:f7b}\includegraphics[width = 0.49\columnwidth]{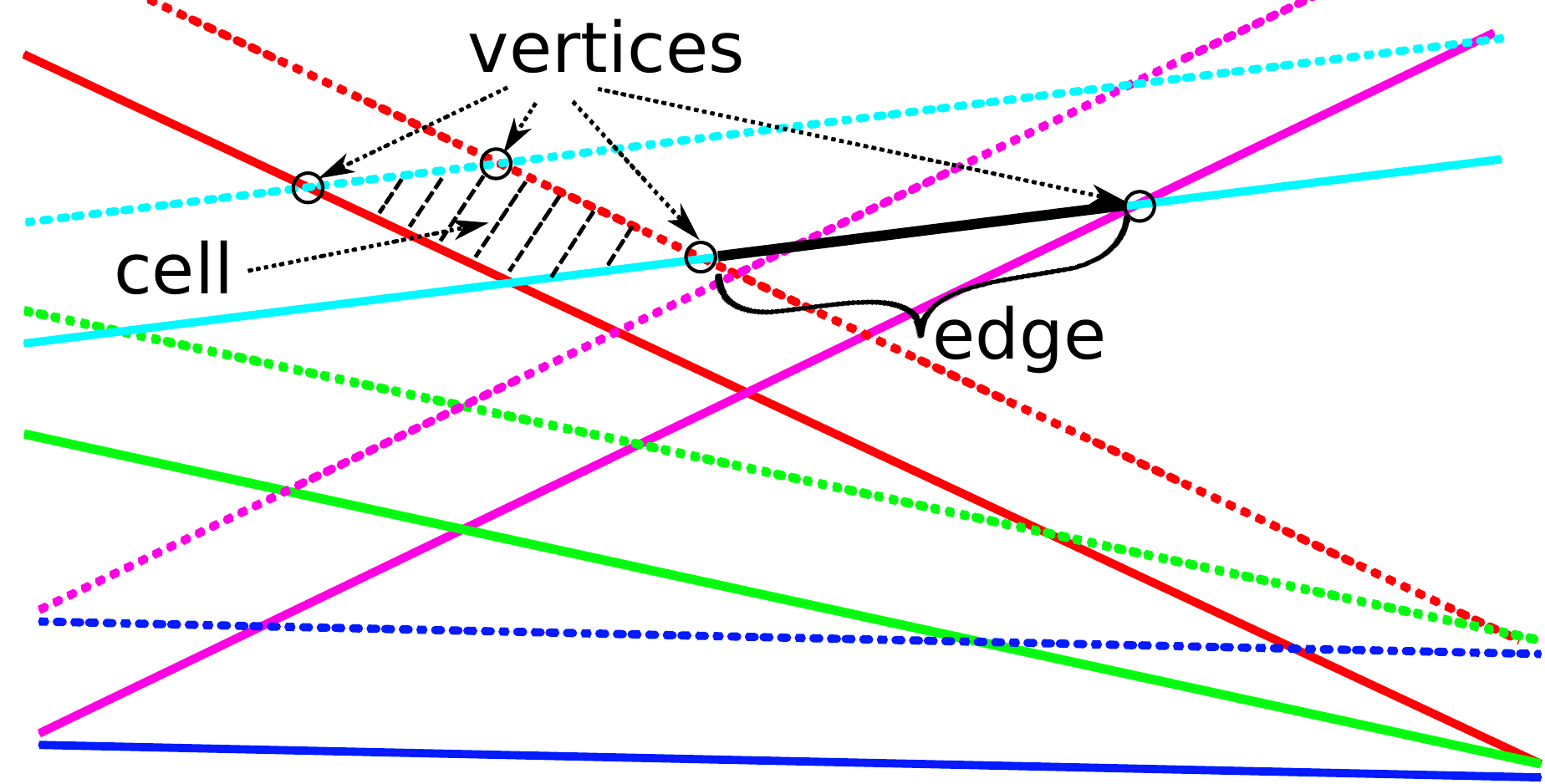}}
			\caption{Point sets $\mathcal{D}'$ and $\mathcal{D}''$ in primal (a) and dual (b) space. In (b), dash and solids lines represent $\bell'_i$ and $\bell''_i$ respectively. In the dual space, a strip is the area between parallel lines $\bell'_i$ and $\bell''_i$, \emph{vertices} are intersections of dual lines, \emph{edges} are line segments that connect vertices, and \emph{cells} are the areas within the intersection of several strips.}
			\label{fig:f7}
		\end{figure}

		To develop useful insights of the feasibility conditions in Definition~\ref{def:feasible} in the dual, we use a construction of Kenmochi et al.~\cite{kenmochi2010efficiently} as follows. Given point set $\cD = \{ \bd_i \}^{N}_{i=1}$, we generate two new point sets $\cD^\prime = \{\bm{d}'_i\}_{i=1}^{N}$ and $\cD'' = \{\bm{d}''_i\}_{i=1}^{N}$, where
		\begin{align}
		\bd'_i = (x_i,y_i-\epsilon_1) \;\;\;\; \text{and} \;\;\;\; \bd''_i = (x_i,y_i+\epsilon_1),
		\end{align}
		i.e., vertically translate $\cD$ down and up by a constant $\epsilon_1$. Mapping the new points to $\cQ$ yields the \emph{arrangement} of lines
		\begin{align}
		\cA = \{ \bell'_i \}^{N}_{i=1} \cup \{ \bell''_i \}^{N}_{i=1},
		\end{align}
		where $\bell'_i = \cF(\bd'_i)$ and $\bell''_i = \cF(\bd''_i)$; see Fig.~\ref{fig:f7}.

		For each $i$, $\bell'_i$ and $\bell''_i$ are parallel lines that are separated by distance $2\epsilon_1$ in $\cQ$. Define the ``strip"
		\begin{align}
		\cS_i = \{\bdelta \in \cQ \mid r(\bdelta,\bell'_i)\le 2\epsilon_1 \;\; \text{and} \;\; r(\bdelta,\bell''_i)\le 2\epsilon_1 \},
		\end{align}
		i.e., the region that lies between $\bell'_i$ and $\bell''_i$. The crucial property is that the points $\bdelta$ in the strip $\cS_i$ satisfy
		\begin{align}
		r(\bdelta,\bell_i) \le \epsilon_1,
		\end{align}
		where $\bell_i = \cF(\bd_i)$, hence, such $\bdelta$'s are dual to lines $\bl = \cF^{-1}(\bdelta)$ in $\cP$ that satisfy
		\begin{align}
		|y_i - \left[ x_i~1 \right]\bl| \le \epsilon_1.
		\end{align}

		The arrangement $\cA$ partitions $\cQ$ into a set of \emph{vertices}, \emph{edges} and \emph{cells}; see Fig.~\ref{fig:f7}. Each cell is formed by the intersection of a number of the strips. Under the the reverse dual mapping, the points $\bdelta$ in a cell are equivalent to all the lines $\bl$ in $\cP$ that enable C2 to be satisfied for a particular subset of points in $\cD$. In effect, $\cA$ partitions $\cD$ into subsets of linear structures $\{ \cD_j \}_{j=1}^M$, where the points in $\cD_j$ satisfy C2 in Definition~\ref{def:feasible}.

		For $N$ lines in ``general position'' on a plane, the number of cells in their arrangement is $\cO(N^2)$~\cite{edelsbrunner1989topologically}. Since there are $2N$ lines in our construction, the number of cells (hence, the number $M$ of linear structures that satisfy C2) is also $\cO(N^2)$.

		To apply the same duality concept for C3 in Definition~\ref{def:feasible}, we simply change the definition of a data point from $\bd_i = (x_i,y_i)$ to $\bd_i = (t_i,x_i)$, and the linear offset from $\epsilon_1$ to $\epsilon_2$. 

		\section{Topological Sweep Method}\label{sec:proposed}

		Given an arrangement of $N$ lines, the classical topological sweep algorithm~\cite{edelsbrunner1989topologically} efficiently explores all cells in the arrangement in $\mathcal{O}(N^2)$ time. Our algorithm for FINDALLTRACKS (Algorithm~\ref{al:overall}) uses our specific version of topological sweep that generates all the linear structures from a point set, based on the dual construction in Sec.~\ref{sec:conddual}. Specifically,
		\begin{itemize}
			\item In Step~\ref{step:topo1}, our topological sweep technique (Sec.~\ref{sec:topo}) is invoked on the input point set $\cD$ to generate all linear structures $\{ \cD_1, \dots, \cD_M \}$ that obey condition C2. An additional function performed by our topological sweep is to enforce condition C1 on the reported linear structures.
			\item In Step~\ref{step:topo2}, our topological sweep method (Sec.~\ref{sec:topo}) is invoked on each $\cD_j$ to further break it down into a set of linear structures $\cT_j$ that also satisfy C3.
		\end{itemize}
		The final output of Algorithm~\ref{al:overall} is the set of all feasible tracks in $\cD$ of length greater than two.

		Sec.~\ref{sec:topo} describes our topological sweep routine in detail, while Sec.~\ref{sec:complexity} discusses the computational cost of the topological sweep routine and Algorithm~\ref{al:overall}.

		\begin{algorithm}[t]\label{al:overall}
			\caption{Proposed method for Problem~\ref{prob:findalltracks} based on topological sweep (Algorithm~\ref{al:TS}).}
			\begin{algorithmic}[1]
				\REQUIRE Time-indexed 2D point set $\mathcal{D} = \{ \bm{d}_i \}^{N}_{i=1}$, positive thresholds $\epsilon_1$ and $\epsilon_2$.
				\STATE $\{ \mathcal{D}_1, \dots, \mathcal{D}_M \} \leftarrow \text{TopoSweep}(\cD,\epsilon_1)$.\label{step:topo1}
				\STATE $\mathcal{T} \leftarrow \emptyset$.
				\FOR{$j = 1,\dots,M$}
				\IF{$|\cD_j|>2$}
				\STATE $\cE_j \leftarrow \{ (t_i,x_i) \}_{i \in \cD_j}$.
				\STATE $\mathcal{T}_j \leftarrow \text{TopoSweep}(\cE_j,\epsilon_2)$.\label{step:topo2}
				\STATE $\mathcal{T} \leftarrow \mathcal{T} \cup \mathcal{T}_j$.
				\ENDIF
				\ENDFOR
				\RETURN $\mathcal{T}$.
			\end{algorithmic}
		\end{algorithm}

		\begin{figure}[ht]
			\centering
			\subfloat[Primal space.]{\label{fig:f8a}\includegraphics[width = 0.49\columnwidth]{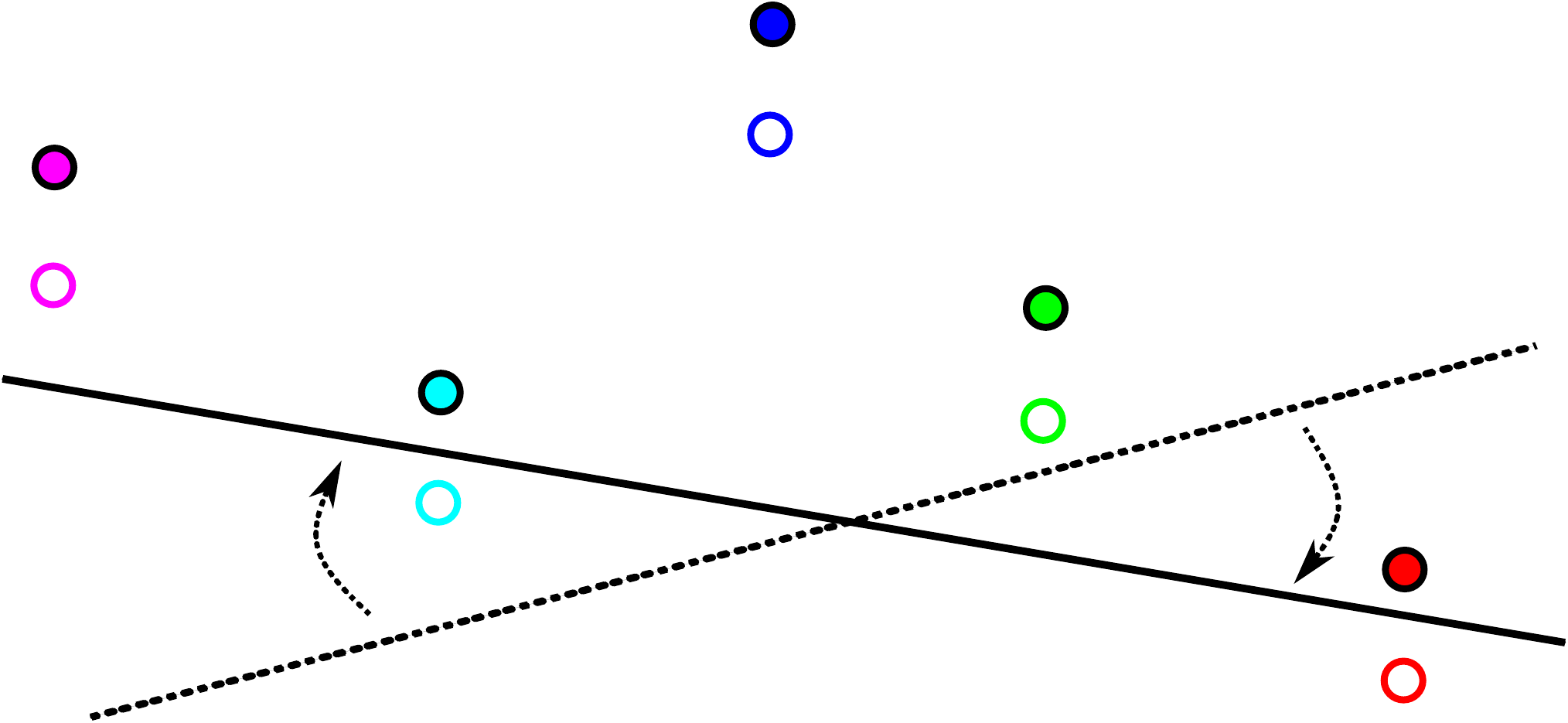}}
			\hfill
			\subfloat[Dual space.]{\label{fig:f8b}\includegraphics[width = 0.49\columnwidth]{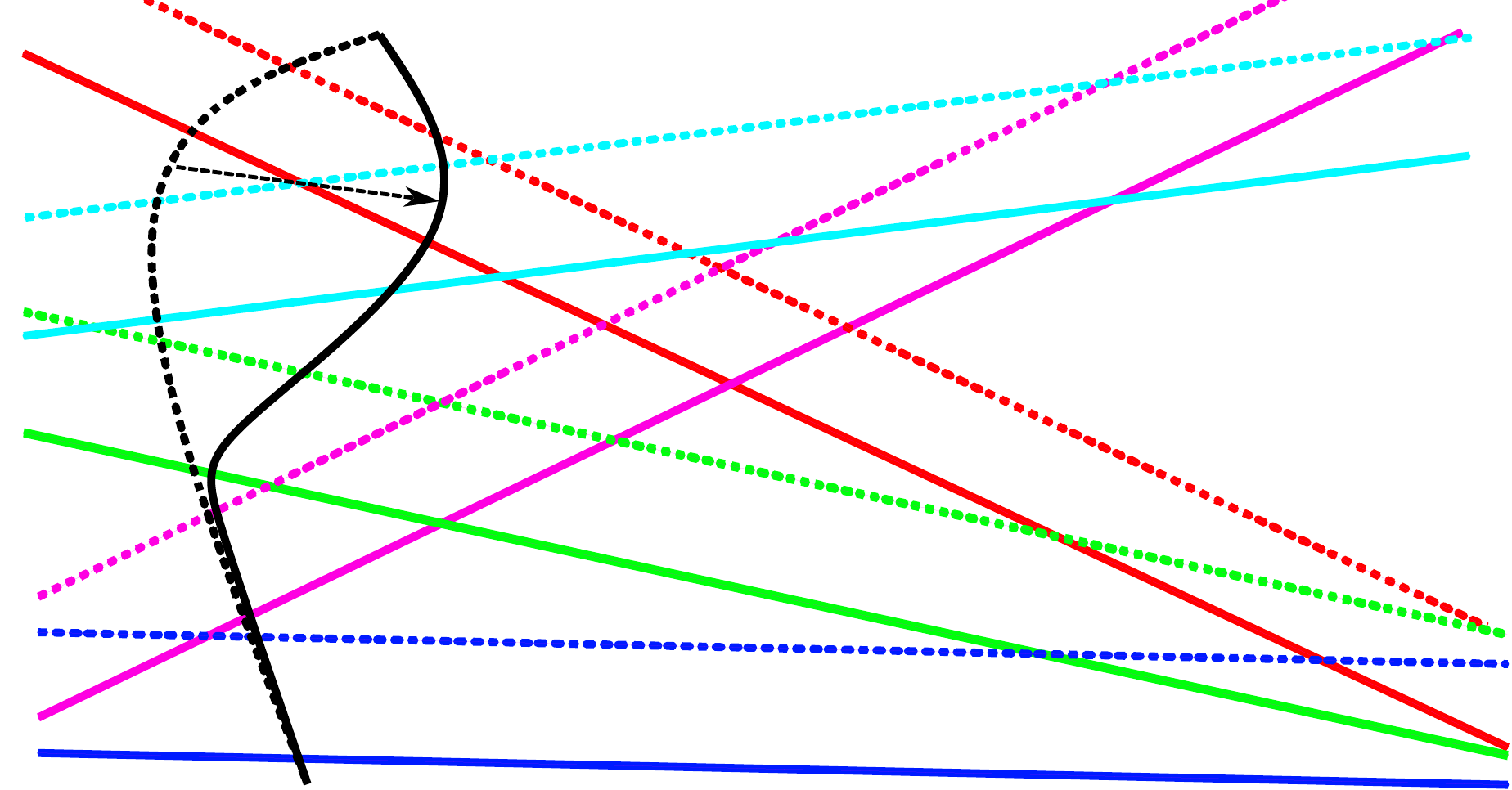}}
			\caption{Elementary step in primal and dual space. (a) Dash and solid line represent the line before and after the elementary step. (b) Dash and solid curve represent topological curve before and after the elementary step.}
			\label{fig:f8}
		\end{figure}

		\subsection{Topological sweep algorithm for track detection}\label{sec:topo}

		The concept of topological sweep is to use a curved line in the plane to traverse a line arrangement, in a way that visits an unseen-before cell at each step; see Fig.~\ref{fig:f8}. In practice, a topological sweep algorithm implements the effects of sweeping and an actual sweep line is not created/maintained. Algorithm~\ref{al:TS} summarizes our version of topological sweep used in Algorithm~\ref{al:overall}. For the original version, see~\cite{edelsbrunner1989topologically}.

		In the following, we first describe the data structures that are required in Algorithm~\ref{al:TS} to achieve the theoretical efficiency of $\cO(N^2)$, before discussing the main algorithmic steps of the technique and the runtime analysis.


		\begin{algorithm}[t]\label{al:TS}
			\caption{Proposed topological sweep algorithm.}
			\begin{algorithmic}[1]
				\REQUIRE $\mathcal{D} = \{\bm{d}_k\}_{i=1}^N$ ,$\epsilon$.
				\STATE $\mathcal{L}$, $\mathcal{L}'$, $\mathcal{L}''$, $\mathcal{Z}$, $\mathcal{C}$, $\mathcal{C_T}$ $\leftarrow$ Initialization($\mathcal{D}$) (Alg.~\ref{al:TS_init}).
				\STATE Initialize $HTU$, $HTL$, $Order$, $Stack$ as in \cite{edelsbrunner1989topologically}
				\WHILE{$Stack$ is not empty}
				\STATE $n \leftarrow$ Pop the line index from $Stack$
				\STATE $p \leftarrow Order(n)$
				\STATE $q \leftarrow Order(n+1)$
				\STATE update $HTU$, $HTL$, $Order$, $Stack$ as in \cite{edelsbrunner1989topologically}
				\STATE $\mathcal{Z}$, $\mathcal{C}$, $\mathcal{C_T}$ $\leftarrow$ update($\mathcal{L}'$, $\mathcal{L}''$, $\mathcal{Z}$, $\mathcal{C}$, $\mathcal{C_T}$, $n$, $p$, $q$)  (Alg.~\ref{al:TS_update})
				\IF{$\mathcal{Z}(n) > 2$ and $\bm{l}_p \in \mathcal{L}'$ and $\bm{l}_q \in \mathcal{L}''$}
				\STATE $M \leftarrow M + 1$
				\STATE $\mathcal{D}_M \leftarrow \{\bm{d}_i \in \mathcal{D} |\mathcal{C}(n,i)=1\}$
				\ENDIF
				\ENDWHILE
				\RETURN Linear structures $\{\mathcal{D}_j\}_{j=1}^M$ .$M \leftarrow 0$
			\end{algorithmic}
		\end{algorithm}



		\subsubsection{Data structures}

		With a collection of dual lines $ \{\bm{\ell}_1,\bm{\ell}_2,\dots,\bm{\ell}_N\}$, plane sweep algorithm~\cite{shamos1976geometric} guarantees to  visit all the line intersections in a specific order with the runtime complexity $\mathcal{O}(N^2log(N))$. Topological sweep algorithm~\cite{edelsbrunner1989topologically,edelsbrunner1990computing,rafalin2002topological} also guarantees to visit all the line intersections with a lower complexity $\mathcal{O}(N^2)$. However, instead of sweeping a straight line in dual space, topological sweep utilizes a curve that cut all the lines exactly once in a specific order. Different from plane sweep, which visits all the intersection from leftmost to rightmost, topological sweep visit each intersection when two consecutive lines have the same right end-point.

		\begin{figure}
			\subfloat[]{\label{fig:f9a}\includegraphics[width = 0.45\linewidth]{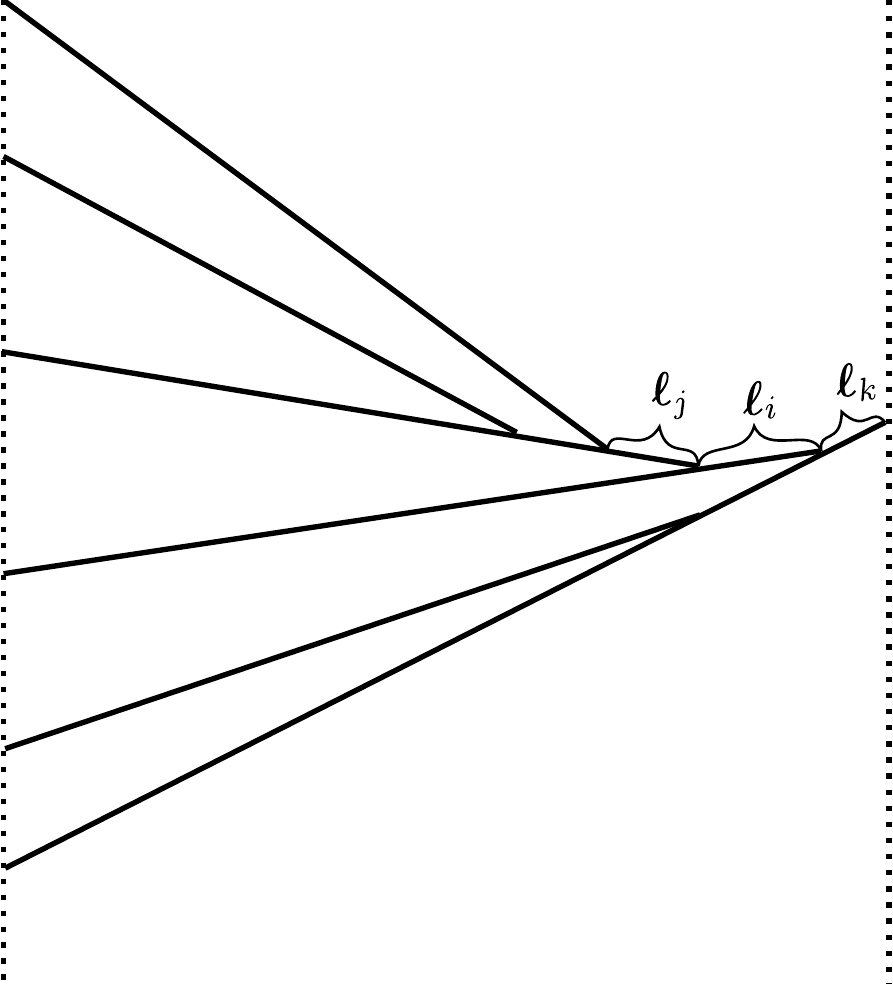}}  \;\;\;\;
			\subfloat[]{\label{fig:f9b}\includegraphics[width = 0.45\linewidth]{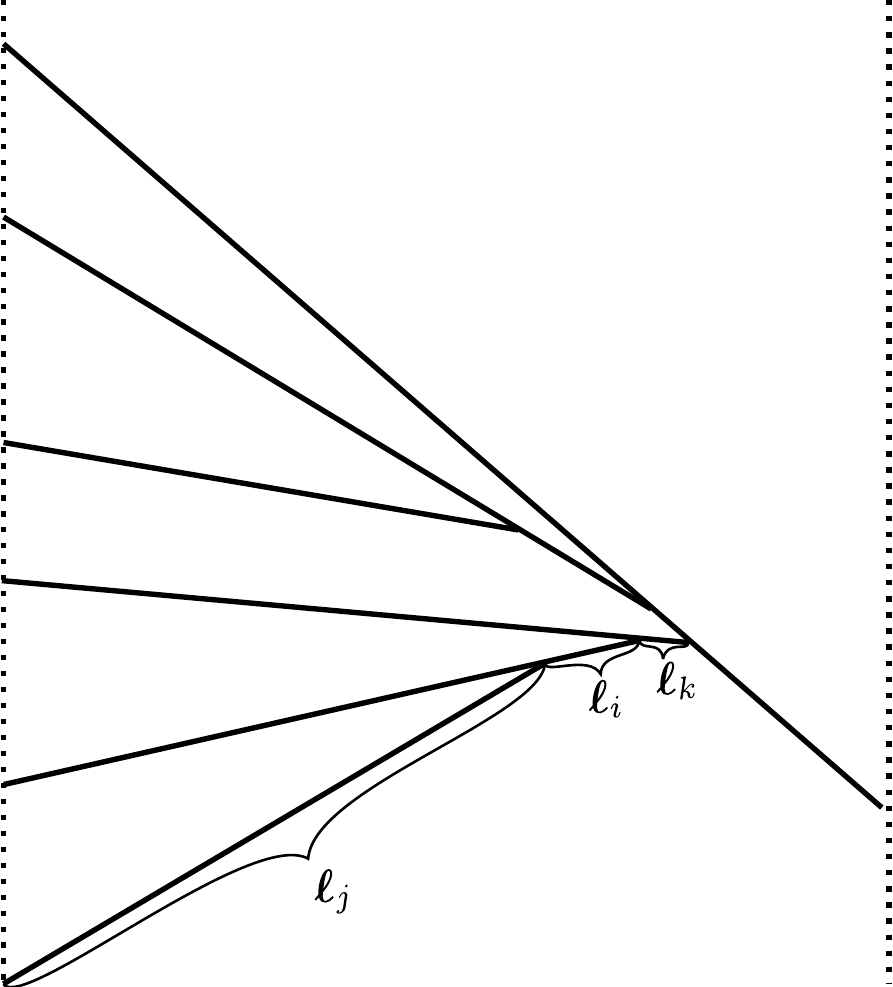}}\\
			\caption{(a): Upper Horizon Tree. Line segments $\bell_j$ and $\bell_k$ delimit the line segment $\bell_i$. Moreover, the slope of $\bell_i$ is smaller than $\bell_k$ and greater than $\bell_j$.  (b): Lower Horizon Tree. The line segment $\bell_i$ is delimited by line segments $\bell_j$ and $\bell_k$. The slope of $\bell_i$ is smaller and $\bell_j$ is greater than $\bell_k$.}
			\label{fig:f9}
		\end{figure}

		Compared to plane sweep, topological sweep utilizes more data structures to record the order of line arrangements, the intersections, and the line segments share the same right end-points. The information recorded in the data structures is further used to perform elementary step. The essential data structure is horizon trees, which are the Upper Horizon Tree $HTU$ and Lower Horizon Tree $HTL$ as illustrated in Fig. (\ref{fig:f9}).  Denote $i,j,k\in\{1,2,\dots,N\} $\par
		\begin{equation}
		\delta(\mathcal{C_T}(n,t)) =
		\begin{cases}
		1, & \text{if}\; \mathcal{C_T}(n,t) = 0\\
		0, & \text{otherwise}
		\end{cases}\label{eq:delta2}
		\end{equation}
		\begin{itemize}
			\item $HTU$: Upper Horizon Tree, which holds the indices of lines segments $(\bm{\ell}_j,\bm{\ell}_k)$ that delimit the line segment $(\bm{\ell}_i$ in the Upper Horizon Tree. The left delimit line segment $\bm{\ell}_j$ always has a smaller slope than $\bm{\ell}_i$ and the right delimit line segment $\bm{\ell}_k$ always has a greater slope than $\bm{\ell}_i$.
			\item $HLU$: Lower Horizon Tree, reverse to Upper Horizon Tree. The left delimit line segment $\bm{\ell}_j$ always has a greater slope than $\bm{\ell}_i$ and the right delimit line segment $\bm{\ell}_k$ always has a smaller slope than $\bm{\ell}_i$.
			\item $Order$: the array holding the sequence of the lines $\mathcal{L}$ cut by the curve from top to bottom.
			\item $Stack$: A stack holds the indexes of lines in $Order$. Each element $i$ in the stack indicates $\bm{\ell}_{Order(i))}$ and $\bm{\ell}_{Order(i+1))}$ has the same right end point.

		\end{itemize}\par

		For details of topological sweep, please refer to~\cite{edelsbrunner1989topologically,rafalin2002topological}. Edelsbrunner and Souvaine~\cite{edelsbrunner1990computing} apply topological sweep to solve the LMS regression line problem with the runtime complexity $\mathcal{O}(N^2)$. More conditions are required for popping and pushing the line intersections $\nu$ into $Stack$ at each elementary step. They also prove $Stack$ never gets empty before visiting all the intersection in their algorithm. They name their algorithm guided topological sweep. Moreover, Shapira and Hassner~\cite{shapira2018fast} use GPU to reduce the processing time of guided topological sweep. However, guided topological sweep algorithm is not suitable for the consensus-based detection problem, sine the outlier ratio is more than $50\%$ and is unknown.

		Besides the data structure that is used to perform the elementary step, one additional array and two metrics are used to record the information of linear structures. Let $\mathcal{C} \in \mathbb{B}^{(2N+1)\times N}$ be a binary matrix, where each row corresponds to one of the regions in the dual space that the topological line is currently visiting, and each column corresponds to one of the pairs of points $\{\bm{d}''_i,\bm{d}'_i\}_{i=1}^N$. Let $\mathcal{C}(n,i)$ denote the entry at the $n$-th row and $i$-th column, which indicates whether a primal line that corresponds to an arbitrary point the in $n$-th region in the dual space stabs the vertical line segment of the $i$-th pair of points $\{\bm{d}''_i,\bm{d}'_i\}$ in the primal space. Let $\mathcal{C_T} \in \mathbb{P}^{(2N+1)\times T}$ be another matrix where its rows have the same meaning as $\mathcal{C}$ and its columns correspond to the $t$ time frames. Let $\mathcal{C_T}(n,t)$ be the entry at row $n$ and column $t$ which counts the number of line segments in time frame $t$ that are stabbed by the $n^{th}$ primal line. Note that $\sum_k \mathcal{C}(n,i) = \sum_t \mathcal{C_T}(n,t)$. For an $n$-th primal line, $\mathcal{C}(n,\cdot)$ records which points are inliers while $\mathcal{C_T}(n,\cdot)$ records how many inliers are from each time frame. However, for tracking problem, each object can only appear once at each time frame. In other words, if there is more than one detection appear on one frame which belong to a single cluster, we only count as one. Finally, the number of inliers of the $n$-th primal line $\mathcal{Z}(n)$ is given by $\mathcal{Z}(n) = \sum_t \mathbb{I}(\mathcal{C_T}(n,t) > 0)$.\par

		\subsubsection{Initialization and update}

		Both Edelsbrunner and Guibas~\cite{edelsbrunner1989topologically} and Kenmochi et al.~\cite{kenmochi2010efficiently} are only searching for line parameters in 2D space. However, our problem is aiming to find all the linear structure with temporal information. Since parallel lines and the temporal information are involved, the initialization and update become more complicated compared to original topological sweep. The initialization and update are summarized in Algorithm~\ref{al:TS_init} and Algorithm~\ref{al:TS_update}. For topological sweep in $t-x$ and $t-y$ subspace, we simply replace $(x,y)$ by $(t,x)$ and $(t,y)$ respectively. our proposed topological sweep algorithm is summarised in Algorithm~\ref{al:TS}.

		\begin{algorithm}[t]\label{al:TS_init}
			\caption{Initialization}
			\begin{algorithmic}[1]
				\REQUIRE $\mathcal{D} = \{\bm{d}_i\}_{i=1}^N$
				\STATE Map each measurement $\bm{d}_i $ to line $\bm{\ell}_i$
				\FOR{ $i \in \{1,\dots,N\}$}
				\STATE $\bm{\ell}'_i \leftarrow (x_i,y_i-\epsilon$);
				\STATE $\bm{\ell}''_i \leftarrow (x_i,y_i+\epsilon$)
				\STATE $\mathcal{L}' \leftarrow \mathcal{L}'\cup l'_i$
				\STATE $\mathcal{L}'' \leftarrow \mathcal{L}'\cup l''_i$
				\ENDFOR
				\STATE $\mathcal{L} \leftarrow \mathcal{L}'' \cup \mathcal{L}'$
				\STATE Sort $\mathcal{L}$ with ascending $x$ value, if two lines are parallel, sort in descending $y$ value
				\STATE $\mathcal{Z}\leftarrow 0$
				\STATE $\mathcal{C}\leftarrow 0$
				\STATE $\mathcal{C_T}\leftarrow 0$
				\FOR{$n \in \{1,\dots,2N\}$}
				\STATE copy $\mathcal{Z}(n)$, $\mathcal{C}(n,\cdot)$ and $\mathcal{C_T}(n,\cdot)$ to $\mathcal{Z}(n+1)$, $\mathcal{C}(n+1,\cdot)$ and $\mathcal{C_T}(n+1,\cdot)$
				\STATE $i \leftarrow$ the corresponding line index $\bm{\ell}_i \in \mathcal{L}''$
				\IF {$\bm{\ell} \in \mathcal{L}''$}
				\STATE $\mathcal{C}(n+1,i)\leftarrow 1$
				\STATE $\mathcal{Z}(n+1)\leftarrow 1$
				\STATE  $\mathcal{C_T}(n+1,t_i)\leftarrow \mathcal{C_T}(n+1,t_i)+1$
				\ELSE
				\STATE $\mathcal{C}(n+1,i)\leftarrow 0$
				\STATE $\mathcal{C_T}(n+1,t_i)\leftarrow \mathcal{C_T}(n+1,t_i)-1$
				\STATE If $\mathcal{C_T}(n+1,t_i) = 0$ then $\mathcal{Z}(n+1)\leftarrow 0$
				\ENDIF
				\ENDFOR
				\RETURN $\mathcal{L}$, $\mathcal{L}'$, $\mathcal{L}''$, $\mathcal{Z}$, $\mathcal{C}$, $\mathcal{C_T}$
			\end{algorithmic}
		\end{algorithm}
		\begin{algorithm}[t]\label{al:TS_update}
			\caption{Update}
			\begin{algorithmic}[1]
				\REQUIRE $\mathcal{L}'$, $\mathcal{L}''$, $\mathcal{Z}$, $\mathcal{C}$, $\mathcal{C_T}$, $n$, $p$, $q$
				\IF{$\bm{l}_p \in \mathcal{L}'$ and $\bm{l}_q \in \mathcal{L}''$ }
				\STATE $\mathcal{Z}(n) \leftarrow \mathcal{Z}(n) + \sum_{j\in \{p,q\}}\delta(\mathcal{C_T}(n,t_j))$ where $\delta$ is given in Eq (\ref{eq:delta2})
				\STATE $\mathcal{C_T}(n,t_p),\mathcal{C_T}(n,t_q) \leftarrow\mathcal{C_T}(n,t_p)+1,\mathcal{C_T}(n,t_q)+1$
				\STATE $\mathcal{C}(n,p),\mathcal{C}(n,q)\leftarrow 1,1$
				\ELSIF {$\bm{l}_p \in \mathcal{L}''$ and $\bm{l}_q \in \mathcal{L}'$ }
				\STATE $\mathcal{C_T}(n,t_p),\mathcal{C_T}(n,t_q) \leftarrow\mathcal{C_T}(n,t_p)-1,\mathcal{C_T}(n,t_q)-1$
				\STATE $\mathcal{C}(n,p),\mathcal{C}(n,q)\leftarrow 0,0$
				\STATE $\mathcal{Z}(n) \leftarrow \mathcal{Z}(n) - \sum_{j\in \{p,q\}}\delta(\mathcal{C_T}(n,t_j))$
				\ELSIF {$\bm{l}_p \in \mathcal{L}'$ and $\bm{l}_q \in \mathcal{L}'$ }
				\STATE $\mathcal{C_T}(n,t_q) \leftarrow\mathcal{C_T}(n,t_q)-1$
				\STATE $\mathcal{C}(n,p),\mathcal{C}(n,q)\leftarrow 1,0$
				\STATE $\mathcal{Z}(n) \leftarrow \mathcal{Z}(n) + \delta(\mathcal{C_T}(n,t_p)) - \delta(\mathcal{C_T}(n,t_q))$
				\STATE $\mathcal{C_T}(n,t_p) \leftarrow \mathcal{C_T}(n,t_p)+1,$
				\ELSIF {$\bm{l}_p \in \mathcal{L}''$ and $\bm{l}_q \in \mathcal{L}''$ }
				\STATE $\mathcal{C_T}(n,t_p) \leftarrow\mathcal{C_T}(n,t_p)-1$
				\STATE $\mathcal{C}(n,p),\mathcal{C}(n,q)\leftarrow 0,1$
				\STATE $\mathcal{Z}(n) \leftarrow \mathcal{Z}(n) + \delta(\mathcal{C_T}(n,t_q)) - \delta(\mathcal{C_T}(n,t_p))$
				\STATE $\mathcal{C_T}(n,t_q) \leftarrow \mathcal{C_T}(n,t_q)+1$
				\ENDIF
				\RETURN $\mathcal{Z}$, $\mathcal{C}$, $\mathcal{C_T}$
			\end{algorithmic}
		\end{algorithm}

		\subsection{Computational cost}\label{sec:complexity}

		The number of cells partitioned by the arrangements is $2N^2 + N +1$ in dual space. The lower bound complexity to visit all the cells is $\mathcal{O}(2N^2 + N +1) = \mathcal{O}(N^2)$. The elementary step performed in the original topological sweep~\cite{edelsbrunner1989topologically} requires only constant time (per step). In our proposed topological sweep (Algorithm~\ref{al:TS}), we update the consensus set within a constant time by adding or removing at most $2$ elements at each elementary step, as stated in the Algorithm~\ref{al:TS_update}. Hence, the effort to execute Algorithm~\ref{al:TS} is $\mathcal{O}(N^2)$.

		Our overall algorithm for FINDALLTRACKS (Algorithm~\ref{al:overall}) invokes Algorithm~\ref{al:TS} in a ``two-tiered" manner: first, Algorithm~\ref{al:TS} is invoked on $\cD$ to generate all linear structures $\{ \cD_1,\dots,\cD_M\}$ that satisfy C1 and C2; this incurs the cost of $\mathcal{O}(N^2)$. Then, on each $\cD_j$, Algorithm~\ref{al:TS} is invoked again to break it into constituent linear structures that also satisfy C3. Assuming that $\cD_j$ is of size $\eta$ on average, the overall cost of Algorithm~\ref{al:overall} is thus $\mathcal{O}(\eta^2N^2)$. In practice, $\eta \ll N$ since the threshold $\epsilon_1$ is typically small relative to the image dimensions, hence, the cost of Algorithm~\ref{al:overall} is close to $\mathcal{O}(N^2)$.

		\section{Results}\label{sec:results}

		We evaluated the accuracy and performance of the proposed algorithm (Algorithm~\ref{al:overall}) on several datasets for GEO object detection. We also compared our algorithm against alternatives that can be applied to the problem; details as follows.

		\subsection{Datasets and preprocessing}

\begin{figure}[ht]\centering
			\subfloat[]{\label{fig:optus_seq}\includegraphics[width = 0.9\columnwidth]{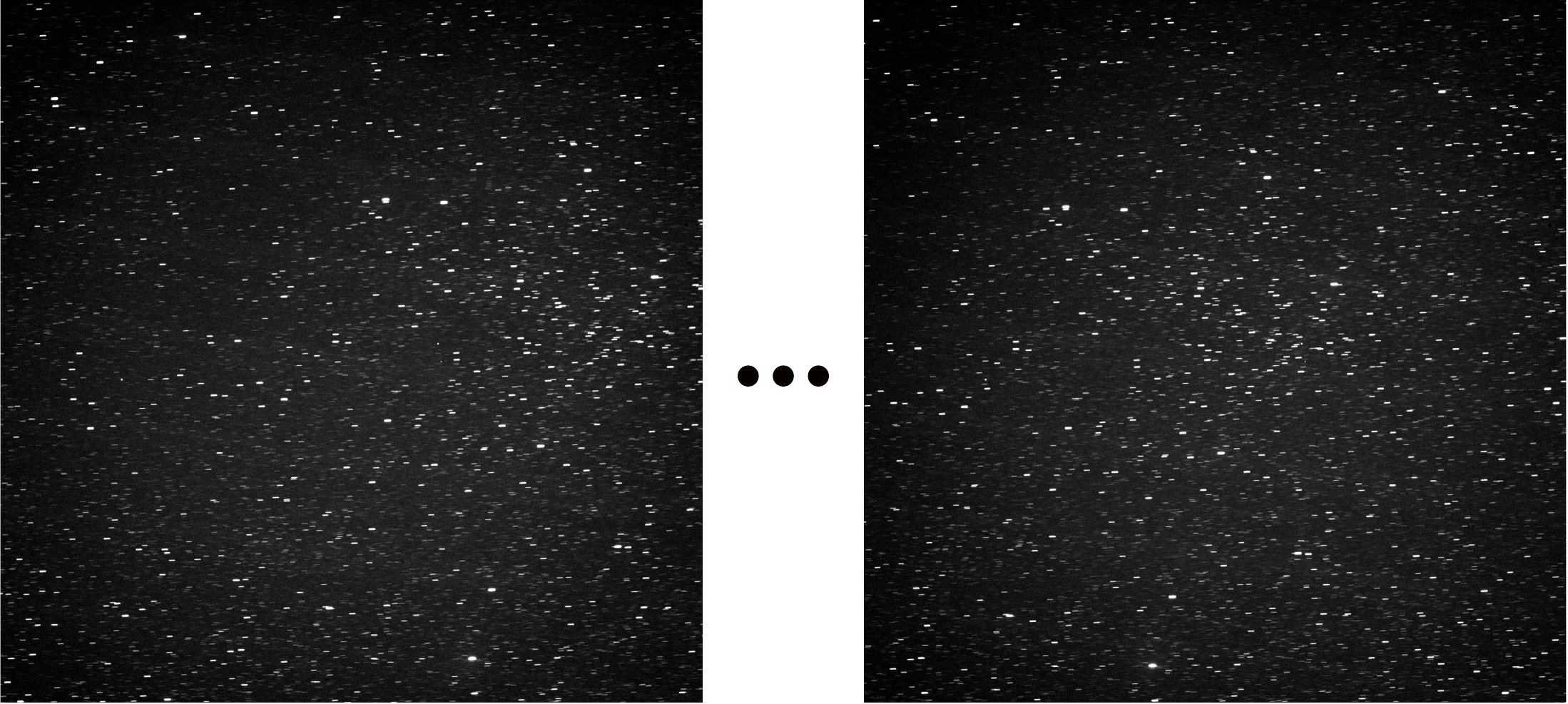}}\\
			\subfloat[]{\label{fig:time_idx_optus}\includegraphics[width = 0.9\columnwidth]{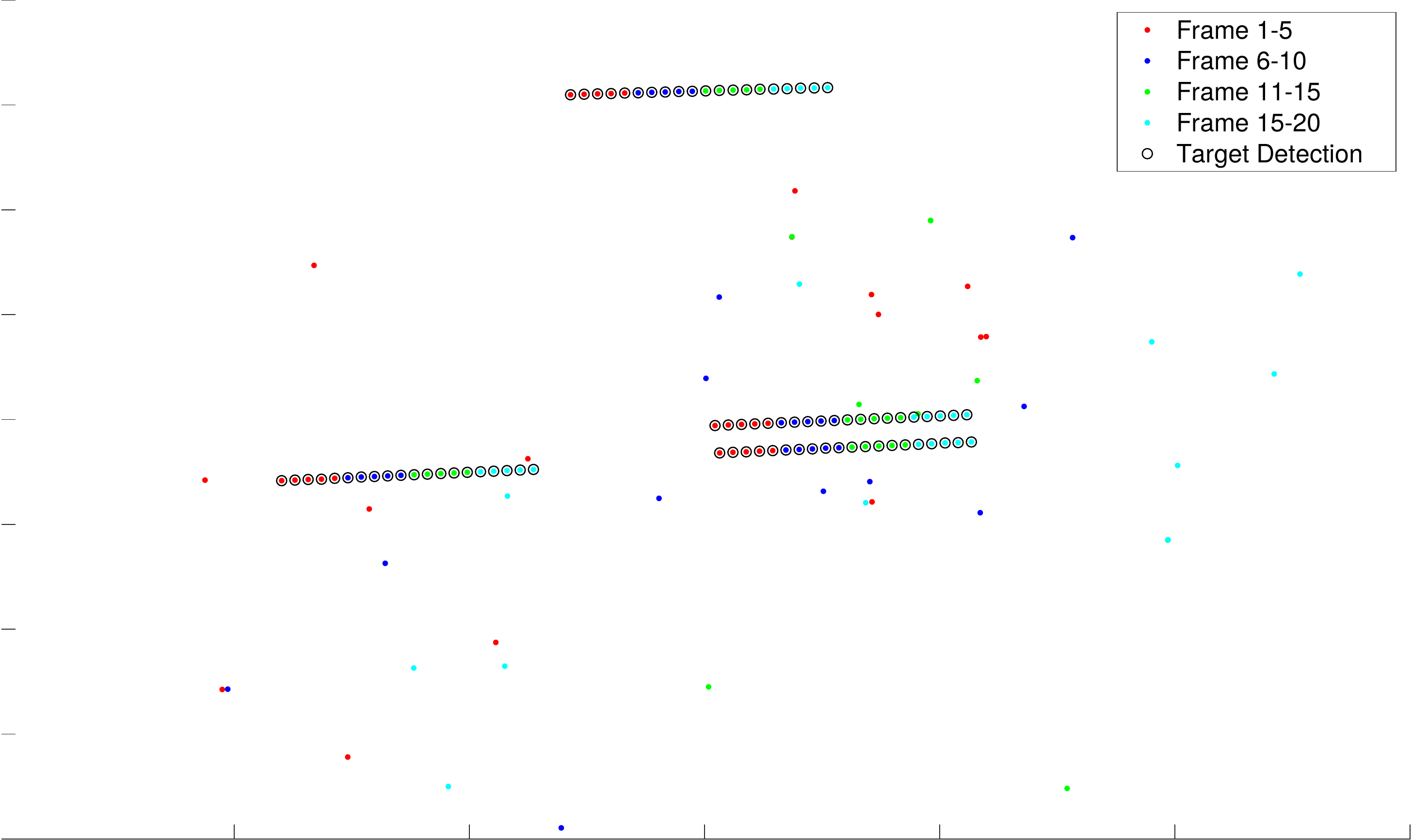}}\\
			\subfloat[]{\label{fig:mosac_optus}\includegraphics[width = 0.9\columnwidth]{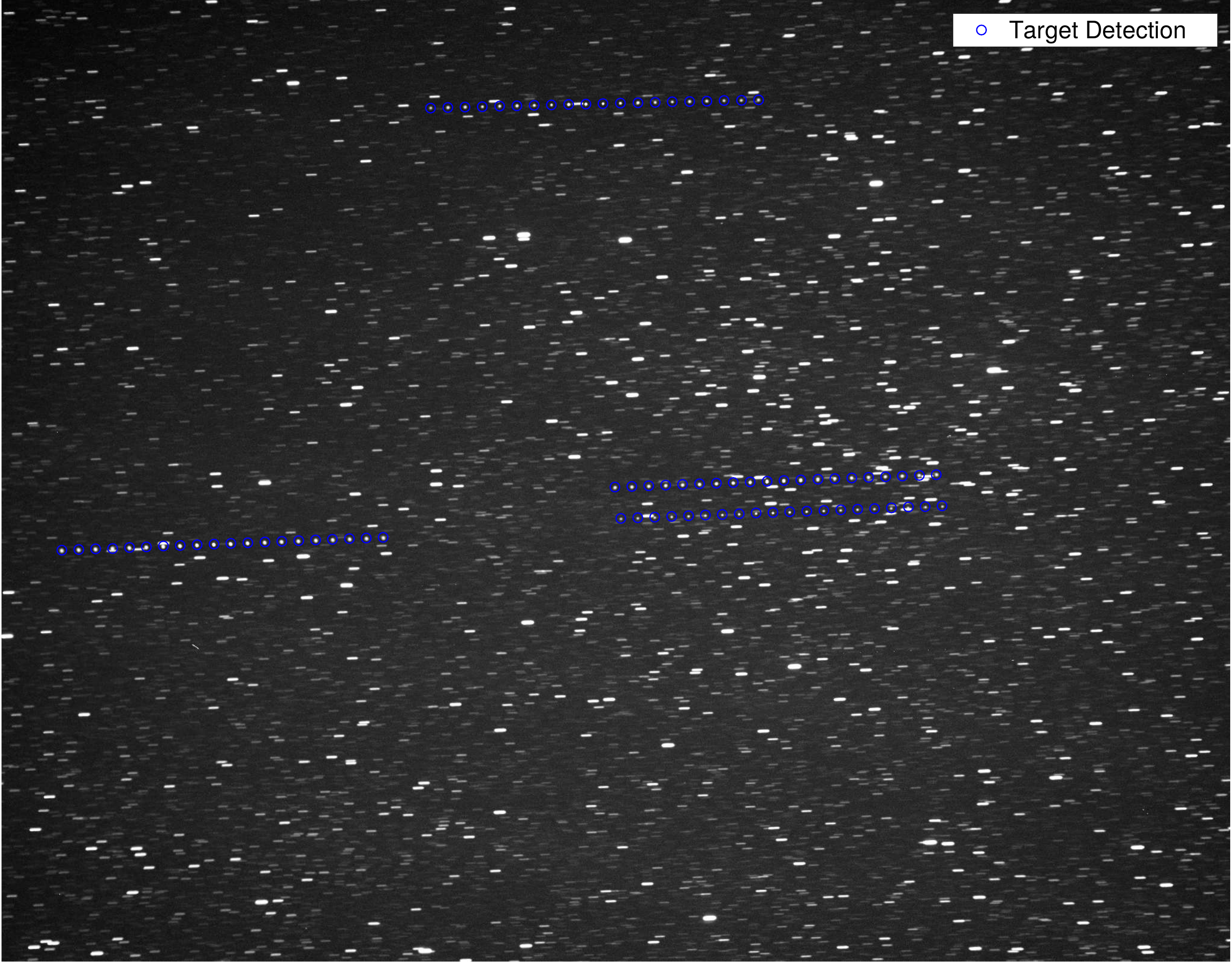}}

			\caption{(a) Sample images from the Optus sequence. (b) Time-indexed 2D point set $\cD$ from a 20-frame subsequence of the Optus sequence. Target objects are circled in black. (c) Target objects plotted on the original images (aligned using the registration parameters estimated during preprocessing).}
			\label{fig:sample_optus}
		\end{figure}

		Two datasets were used in our experiments: Optus and Adelaide-DST. All image sequences in the datasets were captured based on the setting in Sec.~\ref{sec:setting}. The Optus dataset consists of a single image sequence with $111$ frames, with four GEO objects contained therein; see Fig.~\ref{fig:sample_optus} which displays the result on a subsequence of Optus. The Adelaide-DST dataset, previously used in~\cite{do2019robust}, contains image sequences captured across two days; 150826 and 160403. The former contains 27 image sequences, and the latter 30 image sequences. Each sequence has $5$ frames, with varying number of GEO objects (0 to 7). See Fig.~\ref{fig:f1} for a sample sequence from Adelaide-DST; for more details, see~\cite{do2019robust}.

		All images in the datasets are 16-bit grayscale with $2048 \times 2048$ resolution. The preprocessing technique~\cite{do2019robust} (as outlined in Sec.~\ref{sec:preprocessing}) was applied on each sequence to produce a time-indexed 2D point set $\cD$; see Figs.~\ref{fig:f3b} and~\ref{fig:time_idx_optus} for sample preprocessing outputs on both datasets. Note that the Optus sequence is much cleaner than the Adelaide-DST sequences.


		\subsection{Methods}

		We compared Algorithm~\ref{al:overall} (henceforth, ``TS'') against the following alternatives, all of which can be directly executed on $\cD$ for multi-target detection:
		\begin{itemize}[leftmargin=1em]
			\item The classical Hough transform (HT)~\cite{moyer2011multi} was applied for line finding in Algorithm~\ref{alg:baseline} to yield a baseline method.
			\item RANSAC~\cite{fischler1981random} was applied for line finding in Algorithm~\ref{alg:baseline} to yield an approach that resembles~\cite{vsara2013ransacing,do2019robust}.
			\item PMHT~\cite{rago1995comparison,willett2002pmht} was performed on $\cD$ to extract $K$ tracks. Reflecting the lack of prior information on the objects, a track is initialized on each point in the first frame. Upon termination of the algorithm, the $K$ longest tracks were chosen as the overall output.
			\item A variant of PMHT where each track was constrained to be a line (henceforth, ``$K$-lines"~\cite{wang2009k}) was executed on $\cD$ to extract $K$ tracks. $K$-lines allows to inject domain knowledge that each track satisfies conditions C2 and C3, which were not given to standard PMHT. The initilization of $K$-lines follows the initilization of PMHT.
		\end{itemize}
		The correct number of tracks $K$ was given to all methods, hence the aim of each method is to return the best $K$ tracks. Also, we tuned each method for best accuracy.

		As another baseline for the proposed method, plane sweep (PS)~\cite{deberg2008} is used in place of topological sweep in Algorithm~\ref{al:overall}. Note that PS returns exactly the same results as TS, hence this is solely for runtime comparisons.


		\subsection{Evaluation metrics}

		Let $\{ \btau^*_k \}^{K}_{k=1}$ be the ground truth (GT) target tracks for an input sequence, and $\{ \btau^\prime_k \}^{K}_{k=1}$ be the $K$ tracks returned by a particular method for that sequence. The following function
		\begin{align}
		f_{\lambda}(\bd_1,\bd_2) = \mathbb{I}(\| \bd_1 - \bd_2 \|_2 \le \lambda)
		\end{align}
		returns 1 if the two points $\bd_1, \bd_2$ are within a given distance threshold $\lambda$, and 0 otherwise ($\mathbb{I}$ is the indicator function). Then
		\begin{align}
		g(\bd_1, \btau) = \mathbb{I}( \exists \bd_2 \in \btau~\text{such that}~f_{\lambda}(\bd_1,\bd_2) = 1 )
		\end{align}
		returns $1$ if there is a point $\bd_2$ from track $\btau$ that matches point $\bd_1$, and 0 otherwise. We used $\lambda = 3$ (pixels) in our work.

		Following~\cite{do2019robust}, the number of true positives achieved by the method on the sequence is
		\begin{align}
		TP^{(\tau)} = \sum_{k_1=1}^K \mathbb{I}\left( \exists k_2~\text{s.t.}~\sum_{\bd^* \in \btau^*_{k_1}} g(\bd^* ,\btau^\prime_{k_2}) > 0 \right),
		\end{align}
		i.e., the number of GT tracks where at least one point of the track is detected by the method. The number of false negatives is thus the number of GT tracks that were not detected, or
		\begin{align}
		FN^{(\tau)} = K - TP^{(\tau)}.
		\end{align}
		The number of false positives incurred by the method is
		\begin{align}
		FP^{(\tau)} = \sum_{k_2=1}^K \mathbb{I}\left( \sum_{\bd^\prime \in \btau^\prime_{k_2}} g(\bd^\prime ,\btau^*_{k_1}) = 0~~\forall k_1 \right)
		\end{align}
		i.e., the number of tracks returned by the method that does not have any matching points with the GT tracks.

		In this work, we also apply a more fine-grained analysis by computing the metrics at the point level. Specifically, we define another true positive count as
		\begin{align}
		TP^{(d)} = \sum_{k_1=1}^K \sum_{\bd^* \in \btau^*_{k_1}} \mathbb{I}\left( \exists k_2~\text{s.t.}~ g(\bd^* ,\btau^\prime_{k_2}) > 0 \right).
		\end{align}
		In words, $TP^{(d)}$ is the number of points from the GT tracks for which there is at least one matching point from the returned tracks. Then, the number of false negatives is the number of points from the GT tracks for which there are no matches, or
		\begin{align}
		FN^{(d)} = \sum_{k_1=1}^K |\btau^*_{k_1}| - TP^{(d)}.
		\end{align}
		The number of false positives incurred by the method is
		\begin{align}
		FP^{(d)} = \sum_{k_2=1}^K \sum_{\bd^\prime \in \btau^\prime_{k_2}} \mathbb{I}\left(  g(\bd^\prime ,\btau^*_{k_1}) = 0~~\forall k_1 \right)
		\end{align}
		i.e., the number of points from the returned tracks that are not matched with any points from the GT tracks.

		Given the above definitions, we compute the following metrics to evaluate the performance of a method over a dataset:
		\begin{align}
		\begin{gathered}
		Recall^{(z)} = \dfrac{\text{Total $TP^{(z)}$ over all seqs.}}{\text{Total $TP^{(z)} + FN^{(z)}$ over all seqs.}};\\
		Precision^{(z)} = \dfrac{\text{Total $TP^{(z)}$ over all seqs.}}{\text{Total $TP^{(z)} + FP^{(z)}$ over all seqs.}};\\
		F1^{(z)} = \dfrac{2 \cdot Recall^{(z)} \cdot Precision^{(z)}}{Recall^{(z)} + Precision^{(z)}},
		\end{gathered}
		\end{align}
		where $z \in \{ \tau, d \}$. The perfect method would achieve $1$ for all the metrics above. Finally, we also record the runtime of each method on each sequence as a measure of efficiency.

		\subsection{Optus dataset}

		\subsubsection{Accuracy evaluation}

		Given the length of the Optus sequence, we randomly selected $20$-frame subsequences and executed each method on them; see Fig.~\ref{fig:sample_optus} for a sample subsequence. A total of $20$ subsequences were tested, and the average recall, precision and F1 score are summarized in Table.~\ref{tab:optus_acc}, which shows that TS and RANSAC achieved perfect or almost perfect results. However, though RANSAC detected all target tracks, it did not retrieve all target detections, as exhibited by the non-unity $Rec.^{(d)}$. Nonetheless, the results suggest that the Optus dataset is relatively easy; later we will demonstrate clear accuracy gap between TS and RANSAC on the more challenging Adelaide-DST dataset.

		\begin{table}[t]\centering
			\centering
			\caption{Average recall, precision and F1 on 20-frame subsequences of the Optus Dataset. (bold values are the best results)}
			\label{tab:huan_optus}
			\begin{tabular}{|c|c|c|c|}
				\hline
				Method         & $Rec.^{(\tau)}$ & $Prec.^{(\tau)}$ & $F1^{(\tau)}$   \\ \hline
				TS & \textbf{1.0000} & \textbf{1.0000}  &  \textbf{1.0000} \\ \hline
				HT & 0.6875 & 0.9125  & 0.7842 \\ \hline
				RANSAC & \textbf{1.0000} & \textbf{1.0000}  &  \textbf{1.0000}   \\ \hline
				PMHT (ideal init.)            & 0.8938 & 0.8938  & 0.8938   \\ \hline
				$K$-lines (ideal init.) & 0.8750 & 0.8750 & 0.8750 \\ \hline
			\end{tabular}\\
			\vspace{1em}
			\begin{tabular}{|c|c|c|c|}
				\hline
				Method         & $Rec.^{(d)}$ & $Prec.^{(d)}$ & $F1^{(d)}$   \\ \hline
				TS & \textbf{0.9968}  & \textbf{1.0000}  &  \textbf{0.9984} \\ \hline
				HT & 0.6748 & 0.9124  & 0.7758 \\ \hline
				RANSAC       &\textbf{0.9751}  & \textbf{1.0000} &  \textbf{0.9896}  \\ \hline
				PMHT  (ideal init.)             & 0.8912& 0.8912  & 0.8912   \\ \hline
				$K$-lines (ideal init.) & 0.8738 & 0.8738 & 0.8738 \\ \hline
			\end{tabular}
			\label{tab:optus_acc}
		\end{table}

		\subsubsection{Runtime evaluation}

\begin{figure}[ht]\centering
			\subfloat[Runtime of different algorithms versus size $N$ of time-index 2D point sets $\cD$ from subsequences of the Optus dataset.]{\label{fig:time1}\includegraphics[width= 0.99\columnwidth]{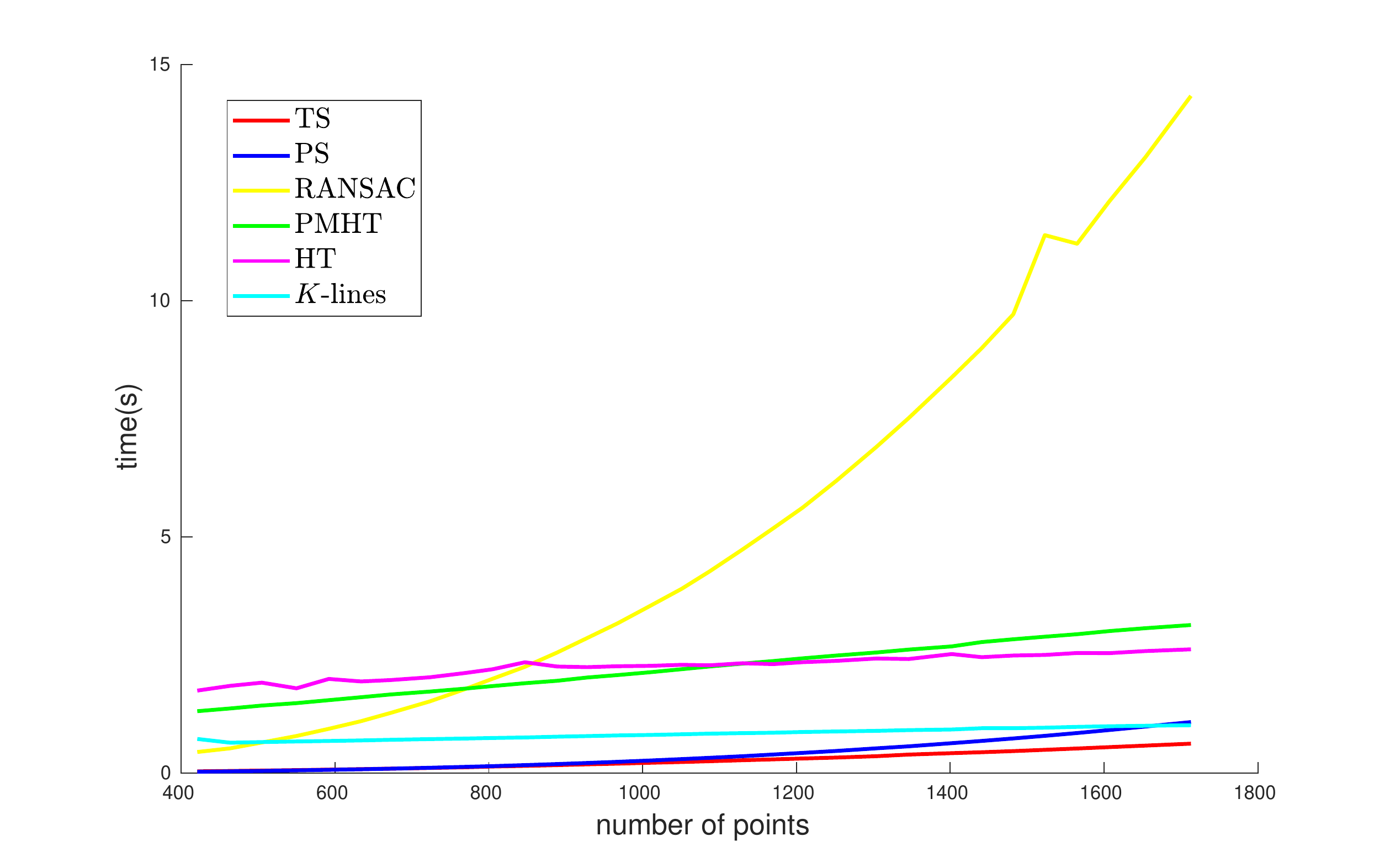}}\\
			\subfloat[Runtime of TS and PS versus size $N$ of time-index 2D point sets $\cD$ from subsequences of the Optus dataset (Legend as Fig.~\ref{fig:time1}).]{\label{fig:time2}\includegraphics[width= 0.99\columnwidth]{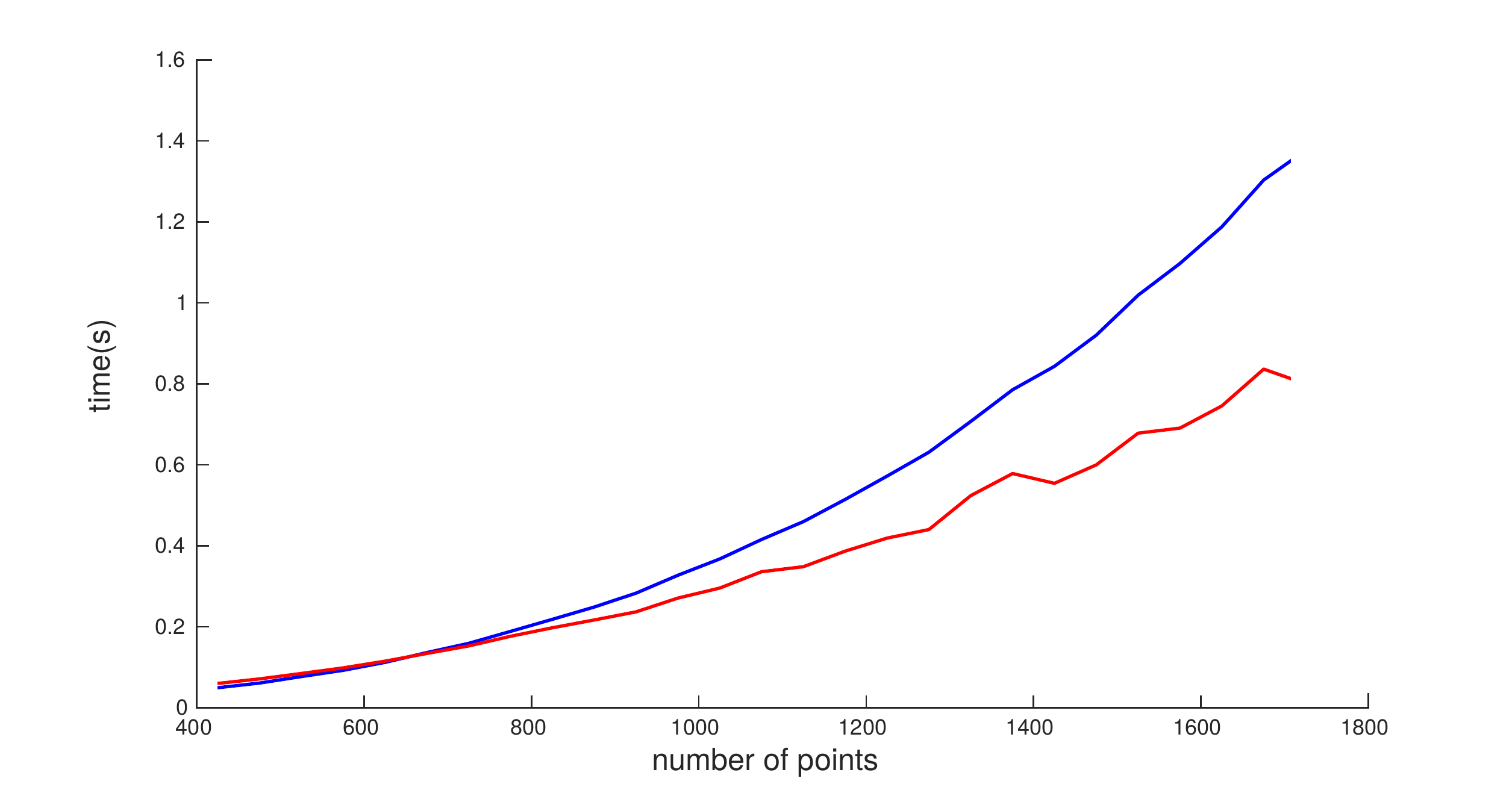}}\\
			\label{fig:Time}
		\end{figure}

		To investigate the computational efficiency of the methods, we executed them on subsequences of the Optus dataset of length $F = 10$ to $F = 40$; via the preprocessing routine, these yielded time-indexed point sets $\cD$ of varying sizes. Since HT, RANSAC, PMHT and $K$-lines are relatively simple algorithms, they were implemented in Matlab, using built-in functions as much as possible. In order to enjoy the computational efficiency of topological sweep, TS and PS was implemented in C++. All the experiments were run on a machine with Intel i5-8600k CPU at 3.6GHz.,

		Since the algorithms were implemented/executed in different environments, a direct runtime comparisons is not entirely fair. However, as we will see soon, differences in the asymptotic runtime of the different algorithms can still be observed clearly. Fig.~\ref{fig:Time} plots the recorded runtime for all methods as a function of size $N$ of time-index 2D point sets $\cD$. The runtimes of TS at $N \approx 800$ and $N \approx 1600$ were $0.1395$ and $0.5885$ seconds respectively, which were only slightly worse than $\mathcal{O}(N^2)$, thus indicating the soundness of the computational analysis in Sec.~\ref{sec:complexity}. The runtime of RANSAC increased rapidly with the number of points; this was because the amount of clutter (outliers) increases with the number of frames (recall that the runtime of RANSAC increases exponentially with the outlier rate~\cite{fischler1981random}).

		\subsection{Adelaide-DST dataset}




		\begin{table}[t]\centering
			\centering
			\caption{Average recall, precision and F1 on the Adelaide-DST dataset. (bold values are the best results)}\label{tab:dst_acc}
			\begin{tabular}{|c|c|c|c|}
				\hline
				\multicolumn{4}{|c|}{Data subset 150826}  \\
				\hline
				Method         & $Rec.^{(\tau)}$ & $Prec.^{(\tau)}$ & $F1^{(\tau)}$   \\ \hline
				TS & \textbf{0.9767} & \textbf{0.9545} &  \textbf{0.9655}  \\ \hline
				HT   & 0.2326 & 0.2326  & 0.2326  \\ \hline
				RANSAC             & \textbf{0.9767} &\textbf{0.9130} & \textbf{0.9395} \\ \hline
				PMHT               & 0.2791 & 0.2791  & 0.2791  \\ \hline
				$K$-lines         & 0.8605 & 0.8605 & 0.8605\\ \hline
			\end{tabular}\\
			\vspace{1em}
			\begin{tabular}{|c|c|c|c|}
				\hline
				\multicolumn{4}{|c|}{Data subset 150826}  \\
				\hline
				Method         & $Rec.^{(d)}$ & $Prec.^{(d)}$ & $F1^{(d)}$   \\ \hline
				TS & \textbf{0.9720} & \textbf{0.9375} &  \textbf{0.9544}  \\ \hline
				HT   & 0.1528 & 0.0968  & 0.1185  \\ \hline
				RANSAC             & \textbf{0.9404} &\textbf{0.8952} & \textbf{0.9172} \\ \hline
				PMHT               & 0.2222 & 0.2222  & 0.2222  \\ \hline
				$K$-lines          & 0.8333 & 0.8333 & 0.8333\\ \hline
			\end{tabular}\\
			\vspace{1em}
			\begin{tabular}{|c|c|c|c|}
				\hline
				\multicolumn{4}{|c|}{Data subset 160403}  \\
				\hline
				Method         & $Rec.^{(\tau)}$ & $Prec.^{(\tau)}$ & $F1^{(\tau)}$   \\ \hline
				TS & \textbf{0.9259} & \textbf{0.8929} &  \textbf{0.9091}  \\ \hline
				HT   & 0.2593 & 0.2500  & 0.2546  \\ \hline
				RANSAC             & \textbf{0.9259} &\textbf{0.6154} & \textbf{0.7394} \\ \hline
				PMHT               & 0.1111 & 0.1111  & 0.1111  \\ \hline
				$K$-lines          & 0.7407 & 0.7407 & 0.7407\\ \hline
			\end{tabular}\\
			\vspace{1em}
			\begin{tabular}{|c|c|c|c|}
				\hline
				\multicolumn{4}{|c|}{Data subset 160403}  \\
				\hline
				Method         & $Rec.^{(d)}$ & $Prec.^{(d)}$ & $F1^{(d)}$   \\ \hline
				TS & \textbf{0.9248} & \textbf{0.8786} &  \textbf{0.9011}  \\ \hline
				HT   & 0.2406 & 0.2500  & 0.2344  \\ \hline
				RANSAC             & \textbf{0.9248} &\textbf{0.6049} & \textbf{0.7314} \\ \hline
				PMHT               & 0.0902 & 0.0902  & 0.0902  \\ \hline
				$K$-lines          & 0.7068 & 0.7068 & 0.7068\\ \hline
			\end{tabular}\\
		\end{table}

		In practical circumstances, it may not be possible to have long input sequences such as the Optus sequence. This is reflected in the Adelaide-DST dataset where each sequence has $5$ frames only, which increases the difficulty of the problem. Moreover, as depicted earlier, there is significantly more noise, imaging artifacts and clutter in the Adelaide-DST dataset.

		Table~\ref{tab:dst_acc} summarizes the average accuracy of all methods on the dataset, separated according to the two collection dates (150826 and 160403). It is clear that none of the methods achieved perfect results; however, TS is clearly the best performing method. The more challenging data also led to a significant drop in accuracy for HT and PMHT; however, the fact that the linear constraints are provided to $K$-lines enabled it to achieve results of acceptable quality. Note also that the accuracy of RANSAC according to the metrics used in~\cite{do2019robust} is very similar to that reported in~\cite{do2019robust}.

		However, the value of $K$, which is the input to PMHT and $K$-lines and PMHt, is unknown in real applications. In contrast, the proposed algorithm can find the tracks without the number of $K$. We manually set a threshold $Tr$ and return all the tracks with length higher than the threshold value $Tr$. For instance, we are given a sequence of 5 frames and find all the tracks with their length higher than $Tr = 3$.

		\section{Conclusion}\label{sec:conclusions}

		We introduced a novel  topological sweep-based approach for multitarget GEO object detection in SSA using ground-based optical observations. Our method is deterministic, efficient, and insensitive to initialization. We compared our algorithms to PMHT, $K$-lines, HT and RANSAC on real-world datasets, which illustrated the superior accuracy and performance of our method. A potential future work is to integrate our algorithm with an Bayesian filter~\cite{salmond2001particle,Mark2005} to solve problems with nonlinear kinematic models.



		\ifCLASSOPTIONcaptionsoff
		\newpage
		\fi

		\nocite{*}
		\bibliographystyle{IEEEtran}
		\bibliography{reference}

		%





	\end{document}